\documentclass[sigconf]{acmart}
\usepackage{hyperref}
\usepackage{makecell}
\usepackage{multirow}
\usepackage{algorithm}
\usepackage{algorithmic}
\usepackage{subcaption}
\usepackage{mathrsfs}

\newcommand{\bGamma}{\boldsymbol{\Gamma}}
\newcommand{\bTheta}{\boldsymbol{\Theta}}
\newcommand{\bLambda}{\boldsymbol{\Lambda}}

\newcommand{\btau}{\boldsymbol{\tau}}
\newcommand{\bnu}{\boldsymbol{\nu}}
\newcommand{\bA}{\textbf{A}}
\newcommand{\bB}{\textbf{B}}
\newcommand{\bD}{\textbf{D}}

\newcommand{\bF}{\textbf{F}}
\newcommand{\bI}{\textbf{I}}
\newcommand{\bL}{\textbf{L}}
\newcommand{\bT}{\textbf{T}}
\newcommand{\bV}{\textbf{V}}
\newcommand{\bW}{\textbf{W}}
\newcommand{\bX}{\textbf{X}}
\newcommand{\bY}{\textbf{Y}}
\newcommand{\bb}{\textbf{b}}

\AtBeginDocument{%
  \providecommand\BibTeX{{%
    \normalfont B\kern-0.5em{\scshape i\kern-0.25em b}\kern-0.8em\TeX}}}

\copyrightyear{2020}
\acmYear{2020}
\setcopyright{acmcopyright}\acmConference[KDD '20]{Proceedings of the 26th ACM SIGKDD Conference on Knowledge Discovery and Data Mining}{August 23--27, 2020}{Virtual Event, CA, USA}
\acmBooktitle{Proceedings of the 26th ACM SIGKDD Conference on Knowledge Discovery and Data Mining (KDD '20), August 23--27, 2020, Virtual Event, CA, USA}
\acmPrice{15.00}
\acmDOI{10.1145/3394486.3403358}
\acmISBN{978-1-4503-7998-4/20/08}

\begin{document}
\fancyhead{}
	\title[Hybrid Spatio-Temporal Graph Convolutional Network]{Hybrid Spatio-Temporal Graph Convolutional Network: Improving Traffic Prediction with Navigation Data}

\author{Rui Dai}
\authornote{Rui Dai and Shenkun Xu contributed equally to this paper.}
\affiliation{%
	\institution{Alibaba Group}
	\city{Beijing}
	\country{China}
}
\email{daima.dr@alibaba-inc.com}

\author{Shenkun Xu}
\authornotemark[1]
\affiliation{%
	\institution{Alibaba Group}
	\city{Beijing}
	\country{China}
}
\email{shenkun.xsk@alibaba-inc.com}

\author{Qian Gu}
\affiliation{%
	\institution{Alibaba Group}
	\city{Beijing}
	\country{China}
}
\email{hedou.gq@alibaba-inc.com}

\author{Chenguang Ji}
\authornote{Chenguang Ji is the corresponding author.}
\affiliation{%
	\institution{Alibaba Group}
	\city{Beijing}
	\country{China}
}
\email{chenguang.jcg@alibaba-inc.com}

\author{Kaikui Liu}
\affiliation{%
	\institution{Alibaba Group}
	\city{Beijing}
	\country{China}
}
\email{damon@alibaba-inc.com}

\renewcommand{\shortauthors}{Rui Dai, et al.}

\begin{abstract}
	Traffic forecasting has recently attracted increasing interest due to the popularity of online navigation services, ridesharing and smart city projects. Owing to the non-stationary nature of road traffic, forecasting accuracy is fundamentally limited by the lack of contextual information. To address this issue, we propose the Hybrid Spatio-Temporal Graph Convolutional Network (H-STGCN), which is able to ``deduce'' future travel time by exploiting the data of upcoming traffic volume. Specifically, we propose an algorithm to acquire the upcoming traffic volume from an online navigation engine. Taking advantage of the piecewise-linear flow-density relationship, a novel transformer structure converts the upcoming volume into its equivalent in travel time. We combine this signal with the commonly-utilized travel-time signal, and then apply graph convolution to capture the spatial dependency. Particularly, we construct a compound adjacency matrix which reflects the innate traffic proximity. We conduct extensive experiments on real-world datasets. The results show that H-STGCN remarkably outperforms state-of-the-art methods in various metrics, especially for the prediction of non-recurring congestion.
\end{abstract}



\maketitle

\section{INTRODUCTION\label{introduction}}
Spatio-temporal forecasting has important applications such as weather prediction, transportation planning, etc. Traffic prediction is one classic example. Successful deployment of advanced technologies such as time-dependent routing \citep{bast2016route}, intelligent traffic light control \citep{guanjiezhengkdd18}, and proactive traffic management \cite{zhang2011data} rely considerably on the robust performance of traffic prediction.

Forecasting traffic (travel time) is a challenging task as a diverse spectrum of events affect travel demand. While daily commute is relatively predictable, events including festivals, casual entertainment activities, and adverse weather conditions are subject to strong stochasticity and hard to foretell. The absence of such contextual information renders the evolution of road traffic non-stationary \cite{tsymbal2004problem}. As a consequence, prior data-driven approaches \cite{lv2015traffic,li2018diffusion,yu2018spatio} that use state variable (travel time) as the main input generally performed suboptimally. Several studies \cite{he2013improving,liao2018deep} incorporated event-relevant features, for example tweet counts or crowd map queries in the model to handle this issue. Its efficacy, however, is restricted to neighborhood of hot spots.

\begin{figure}
	\centering
	\includegraphics[width=1\linewidth]{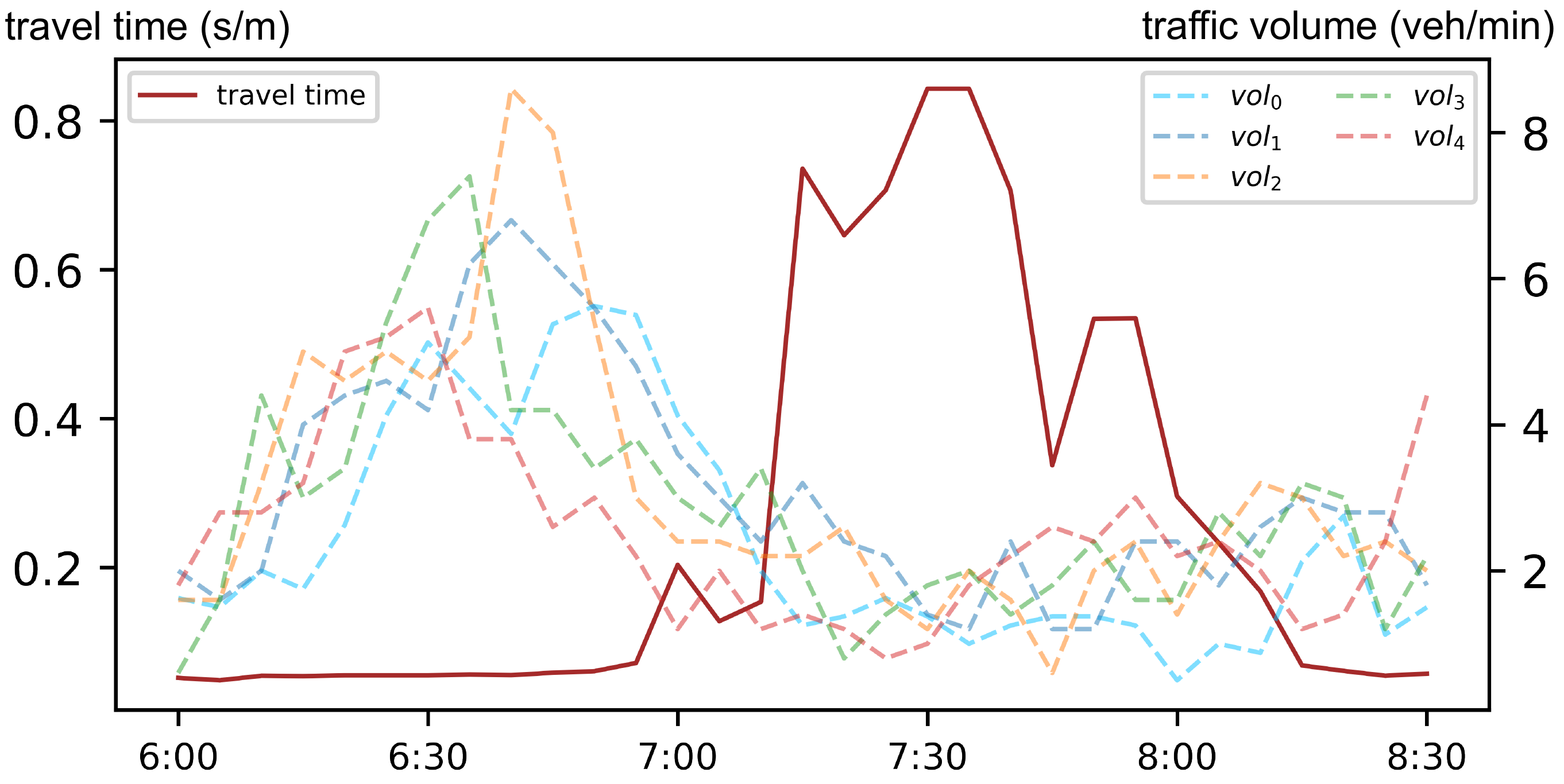}
	\caption{The travel time and intended traffic volume of a road segment in Beijing during morning rush hour on October 28, 2019. $ {vol}_f $ represents the intended traffic volume at a time slot $ f $-step ahead, acquired from the navigation data. The quick rise of intended traffic volume indicates potential traffic congestion.}
	\label{fig:illustration-of-navi}
\end{figure}

To overcome this problem, we augment machine learning models with intended traffic flow acquired from an online navigation engine. Presently, navigation services including smart route recommendations, audio maneuver guidance, etc., are substantially relied upon by drivers in their daily travel. For instance, according to a third-party\footnote{https://www.questmobile.com.cn/en} report, Amap, the top-tier LBS-service provider in China, served more than 115 million users on the National Day of the People's Republic in 2018. The vast number of planned routes offered by a navigation engine comprehensively reflect live travel demand and provide even greater detail than numerous event-level features. More specifically, aggregating the planned routes produces intended traffic volume, which in turn offers strong clues as to future travel time. Figure \ref{fig:illustration-of-navi} illustrates this process.

\begin{figure}[h]
	\centering
	\begin{minipage}[t]{1\linewidth}
		\begin{subfigure}[t]{1\textwidth}
			\centering
			\includegraphics[width=1.0\linewidth]{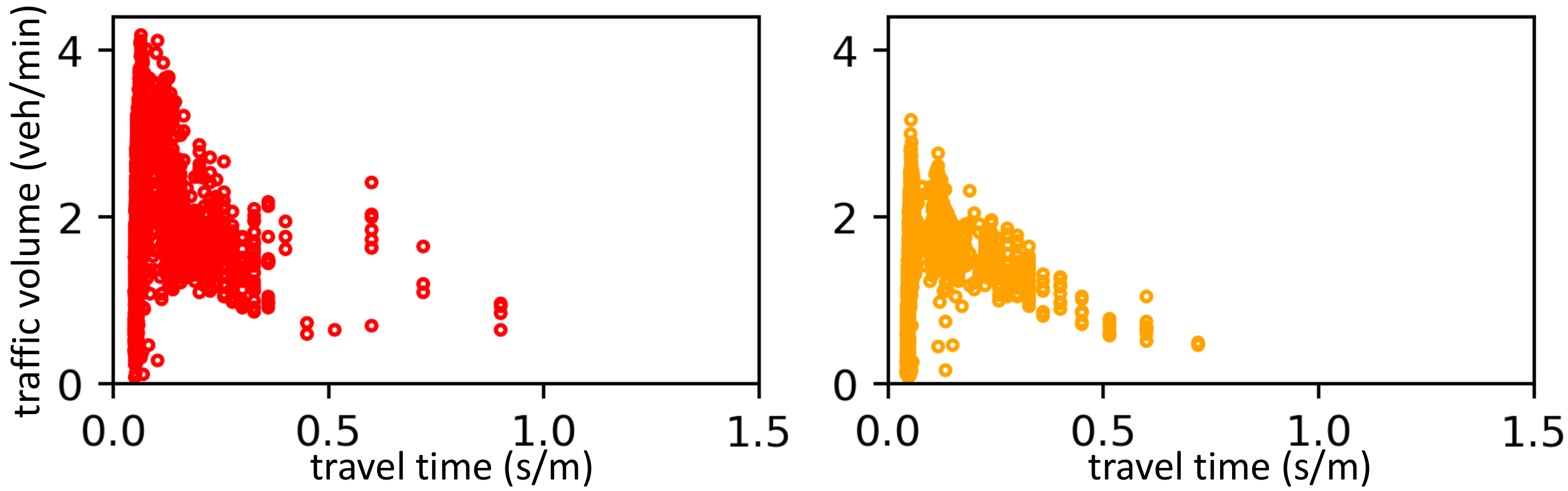}
			\caption{Flow-time curve of expressways}
		\end{subfigure}
		\vfill
		\begin{subfigure}[t]{1\textwidth}
			\centering
			\includegraphics[width=1.0\linewidth]{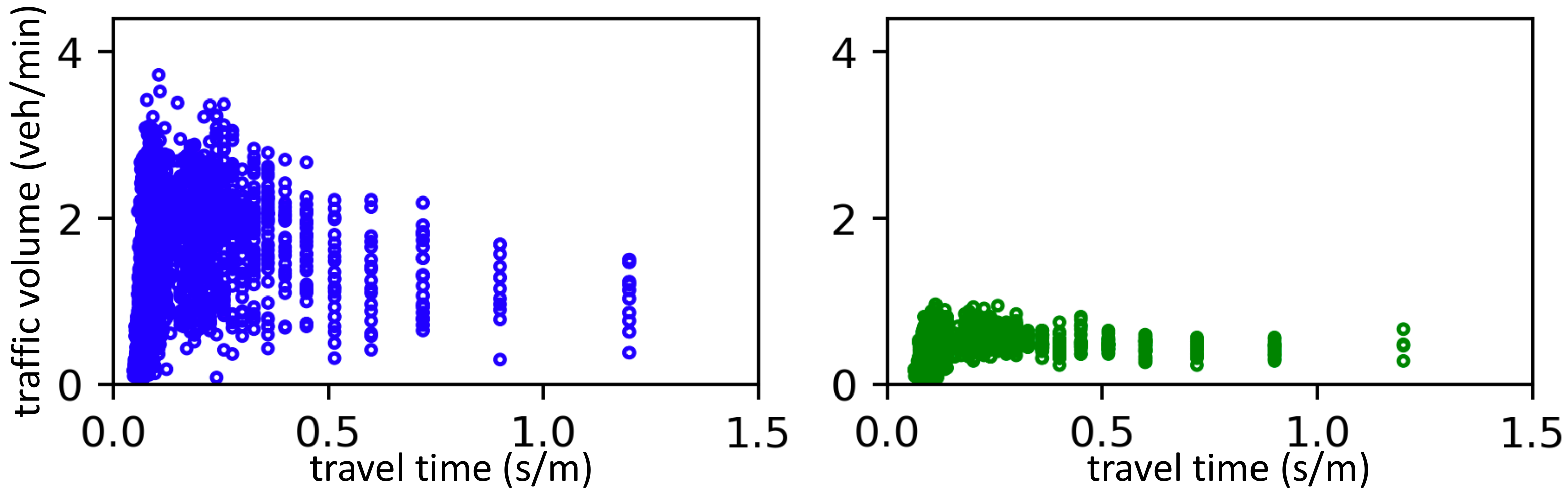}
			\caption{Flow-time curve of major roads}
		\end{subfigure}
	\end{minipage} 
	\caption{Flow-time curve of four different road segments.}
\label{fig:flow-time}
\end{figure}

To integrate this heterogeneous modality into a travel time forecasting model, we design a novel domain transformer to convert traffic volume into its equivalent in travel time. Traffic flow theory \citep{hoogendoorn2001state} establishes that traffic flow and the vehicle density of a road segment satisfies a universal triangular relationship, and specifics of the flow-density diagram such as peak capacity is segment-varying. Figure \ref{fig:flow-time} depicts several real-world examples. To utilize this knowledge, we engineer a flow-to-time transformer with two cascaded mapping components, which are separately responsible for capturing the shared geometric shape and segment-specific characteristics.

Furthermore, owing to the non-Euclidean spatial dependency of road traffic, we adopt graph convolution to extract the shared pattern, and propose a novel adjacency matrix that better reflects innate traffic proximity. In prior scholarship \citep{li2018diffusion}, the adjacency matrix generally assumed a node proximity with simple exponential distance-decay, which does not comport with on-the-ground realities. For example, congestion on a major thoroughfare rarely propagates to an intersecting private road where only authorized person can travel, even though the two segments are contiguous. To solve this issue, we propose a compound adjacency matrix which, in addition to the aforementioned spatial attenuation term, incorporates the covariance matrix of road segment travel time.

As a coherent combination of the above-proposed techniques, we develop a novel multi-modal learning architecture for traffic forecasting: the Hybrid Spatio-Temporal Graph Convolutional Network (H-STGCN). In H-STGCN, the transformer first converts intended traffic volume acquired from Amap into its equivalent in travel time. Then shared convolution is applied on individual segments along the temporal dimension to extract high-level patterns. Graph convolution with the compound adjacency matrix further processes the concatenated temporal signal to capture the intrinsic traffic dynamics. The integrated structure is trained end-to-end and capable of foreseeing future congestion based on upcoming traffic flux. We evaluate the proposed model using real-world datasets. Extensive experiments demonstrate that our model has shown remarkable improvements over various state-of-the-art benchmarks.

To summarize, the primary contributions of the paper are as follows:
\begin{itemize}
	\item We propose to leverage the data of intended traffic flow in a machine-learning model for travel time forecasting. This approach combines the strengths of the recently emerged data-driven approach and the traditional traffic simulation approach \cite{burghout2004hybrid}.
	\item We design the domain transformer to integrate the heterogeneous modality of traffic flow. This universal coupler naturally adapts to all neural-network based architectures for travel time forecasting.
	\item We propose the compound adjacency matrix, which encodes innate traffic proximity.
	\item We construct H-STGCN, a multi-modal learning architecture that significantly outperforms state-of-the-art benchmarks in real world datasets.
\end{itemize}

The rest of the paper is organized as follows. Section \ref{preliminaries} outlines the preliminary concepts and formulates the traffic prediction problem. Section \ref{methodologies} details the structure of the proposed H-STGCN. Section \ref{experiments} describes the experimental results. Related works are reviewed in Section \ref{related work}. Finally, Section \ref{conclusion} concludes the paper.

\section{PRELIMINARIES\label{preliminaries}}
In this section, we provide definitions and outline the forecasting problem. Given a regional network, intersections split it into $ n $ directional road segments. We further split time into 5-minute intervals, and denote the time range of training set by $ [0, S_{\text{train}}) $, test set by $ [S_{\text{train}}, S_{\text{train}} + S_{\text{test}})$. We format the data as a tensor $ \bX \in \mathbb{R}^{n \times (S_{\text{train}} + S_{\text{test}}) \times C^{(\text{in})}} $, where $ C^{(\text{in})} $ is the number of input features.

\textbf{Travel Time / Traffic Volume}. 
Travel time $ \tau_{i, t} $ is defined as the average traversing time (per unit length) on segment $ s_i $ over time slot $ t $. Similarly, traffic volume $ v_{i, t} $ denotes the number of vehicles entering segment $ s_i $ within time slot $ t $.

\textbf{Ideal Future Volume}.
Given a time slot $t_0$, ideal future volume $ \nu_{i, t_0, f} $ ($ f \ge 0 $) is the counterpart of $ v_{i, t_0 + f} $ under two conceptual assumptions: 1) only vehicles using a navigation service at $t_0$ are considered; 2) each vehicle follows the exact planned path and travels at a speed consistent with ETA (estimated time of arrival).

\textbf{Historical Average (HA)}.
Let $ L $ denote the number of time slots in a week. Then the historical average of variable $ \omega_{i, t} $ (ideal future volume or travel time) is given by 
\begin{align}
	\omega^{(h)}_{i, t} = \frac{1}{W}\sum_{r \equiv t \pmod{L}, r \neq t, r \in [0, S_{\text{train}})}{\omega_{i, r}} ,
\end{align}
where $ W $ is the number of weeks in the training set.

\textbf{Traffic Forecasting}.
Given time $ t $ and all available data, traffic forecasting aims to predict future travel time for the whole network. More specifically, provided the sequence of previous traffic features $ \{ \bX_{:, t-P+1, :}, \ldots, \bX_{:, t, :} \} $, model $ \mathscr{H} $ estimates travel time for the next few slots $\mathscr{H}(\bX_{:, t-P+1, :}, \ldots, \bX_{:, t, :}) = \{ \hat{\btau}_{:, t+1}, \ldots, \hat{\btau}_{:, t+F} \} $, where $ P $ denotes the length of input time series, and $ F $ the forecasting horizon.

\section{METHODOLOGY\label{methodologies}}
\subsection{Overall Architecture}

\begin{figure}[bt!]
	\centering
	\includegraphics[width=0.93\linewidth]{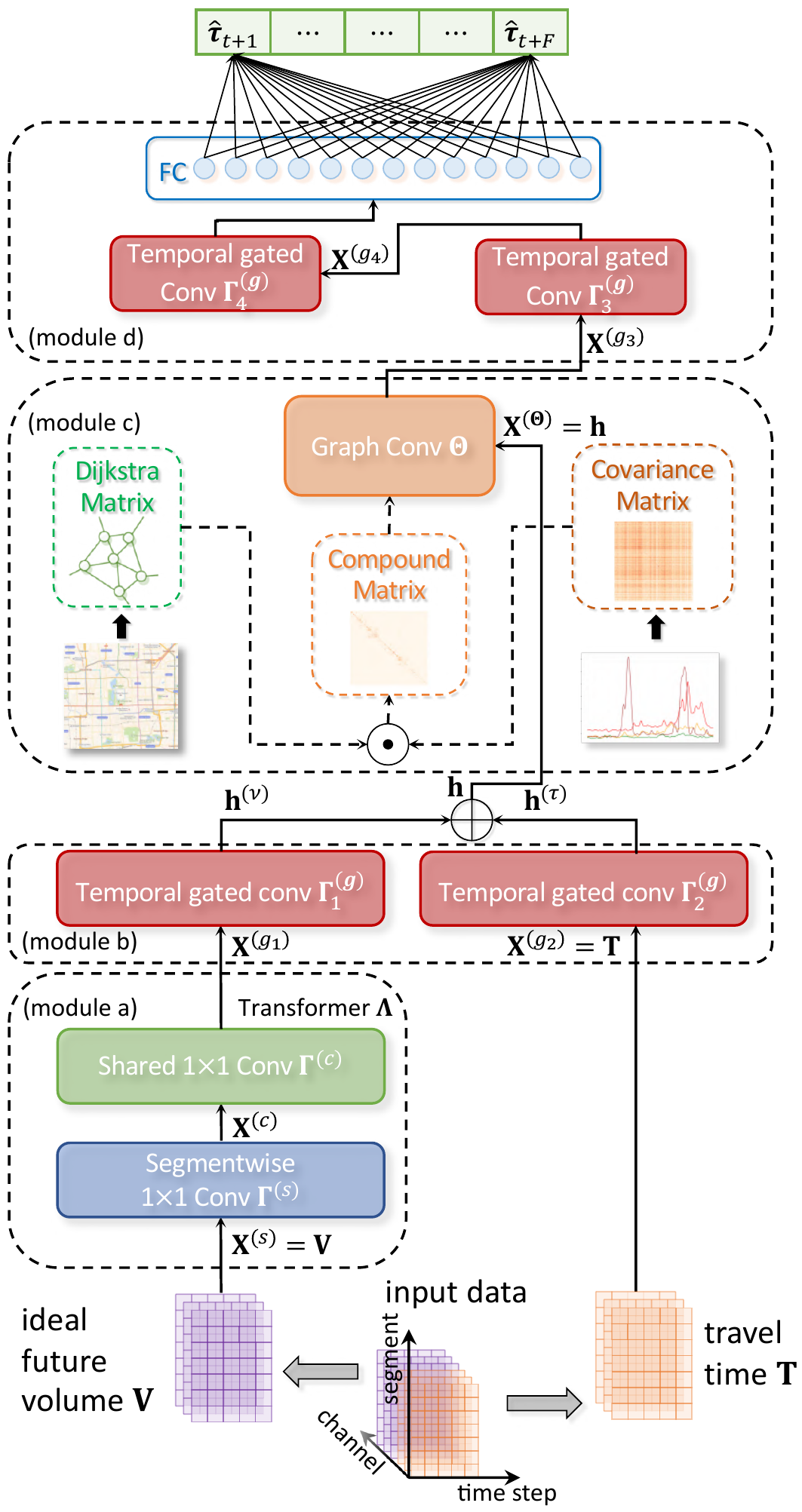}
	\caption{Architecture of H-STGCN.}
	\label{fig:model-detail}
\end{figure}

In this section, we describe the overall architecture of H-STGCN, as illustrated in Figure \ref{fig:model-detail}. The model input consists of two feature tensors, the ideal-future-volume tensor $ \bV $ and travel-time tensor $ \bT $. Specifically, both $ \bV $ and $ \bT $ have three dimensions: the spatial dimension, temporal dimension, and channel dimension, which corresponds respectively to the road segments, previous time slots utilized, and features. A domain transformer (module a) first converts  each element of $ \bV $ into its equivalent in travel time, outputting the so-called future-travel-time tensor $ \bX^{(g_1)} $. Then separate gated convolutions along the temporal dimension (module b) are applied on $ \bX^{(g_1)} $ and $\bT$ to extract the high-level temporal patterns. Treating each segment as a node, a graph convolution with compound adjacency matrix (module c) processes the mixed signal $ \textbf{h} = \textbf{h}^{\nu} \oplus \textbf{h}^{\tau} $ to capture the interaction mechanism between traffic volume and travel time (``$ \oplus $'' stands for the concatenation operator). Next, two additional gated convolutions are applied sequentially to further enlarge the temporal receptive field.  Finally, a fully connected (FC) layer outputs the forecasting results. We elaborate each of the modules in subsequent sections.

\subsection{Model Input and Data Processing}
To forecast future traffic states, H-STGCN uses an input tensor $ \bX $ with features from all $ P $ previous time slots. Each slice of $ \bX $ that corresponds to a single time slot $ t $ ($ \le t_0 $) further comprises two categories of features: ideal future volume and travel time. These features and associated data processing techniques are described as follows.

\begin{figure}
	\centering
	\includegraphics[width=0.8\linewidth]{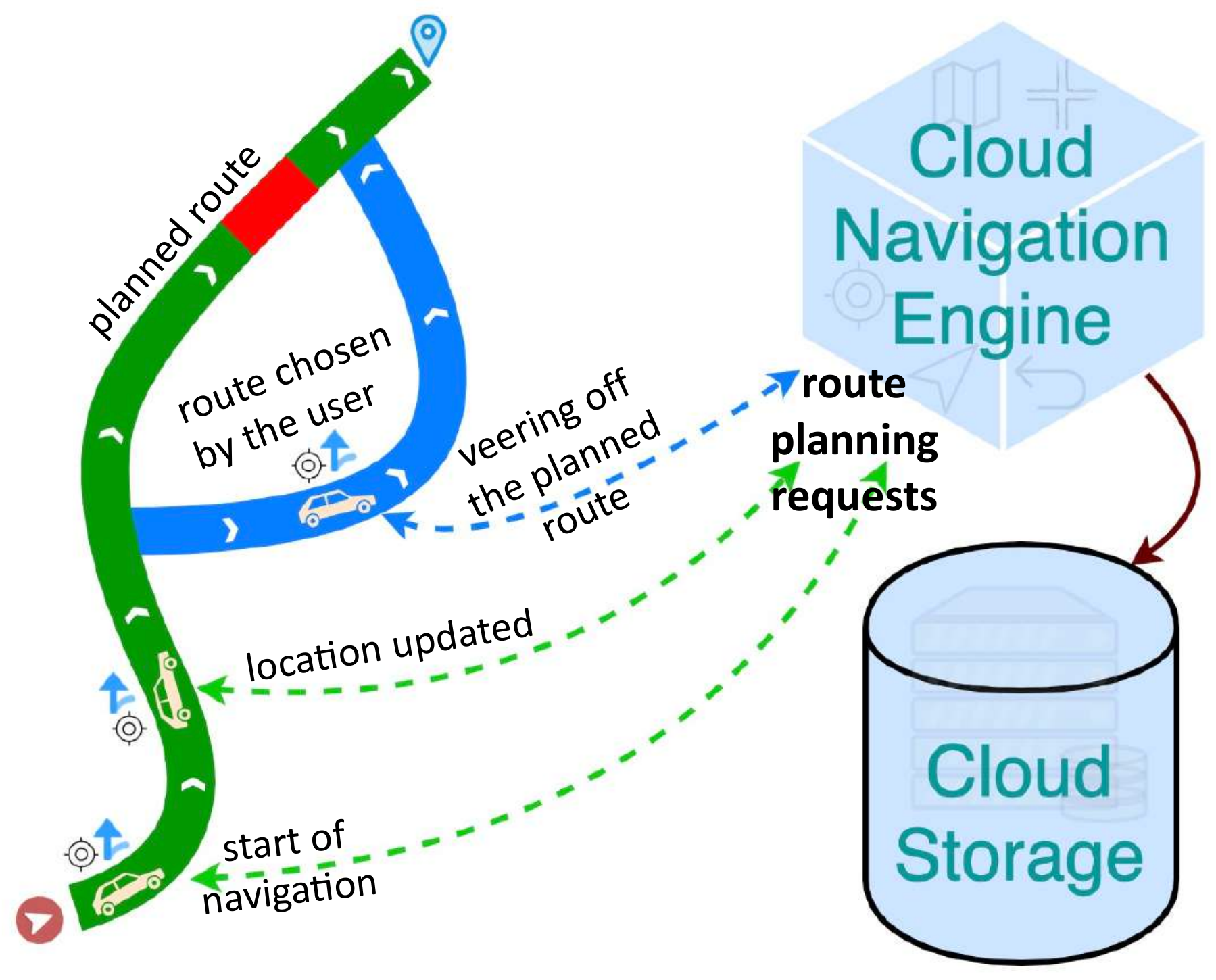}
	\caption{Architecture of Amap's navigation system.}
	\label{fig:navi-engine}
\end{figure}

\textbf{Ideal Future Volume}.
As an approximation of the unavailable actual future traffic volume, the ideal future volume $ \nu_{i, t_0, f} $ defined in Section \ref{preliminaries} can be acquired from an online navigation engine. To employ this feature, we use data from Amap, a leading LBS solution provider in China with over 700 millions users\footnote{https://www.iresearch.com.cn/}. The architecture of Amap’s navigation system is depicted in Figure \ref{fig:navi-engine}. During navigation, a vehicle synchronizes its location with the cloud server every second to ensure the timely detection of potential deviation and follow-up rerouting. Meanwhile, to keep users posted of the latest traffic condition, cloud servers update the ETA in a near-real-time fashion. In this way, the navigation engine is able to collect the planned path and live ETA of a vehicle, with at most a one-second delay when a detour happens.

Original data acquired from Amap is formally organized as
\begin{equation}
	\mathscr{L} = \{(r, \{(\rho_{r, l}, \delta_{r, l}, \psi_{r}) | l \in [0, M_{r}) \} | r \in [0, N_{\mathscr{L}})\} ,
\end{equation}
where $ r $ is the navigation identifier, $ \psi_{r} $ is the launch time of navigation $ r $, $ \rho_{r, l} $ denotes the $ l $-th road segment along the planned route, $ \delta_{r, l} $ is the estimated time to arrive at  $ \rho_{r, l} $, $ M_{r} $ is the total number of road segments on the route, and $ N_{\mathscr{L}} $ is the total number of navigation processes. Specifically, routes in Amap are planned with Dijkstra-like algorithms \citep{bast2016route}, and ETA is forecasted using a machine learning model inferred from historical trajectories. Each record in $ \mathscr{L} $ corresponds exactly to one planned route. Algorithm \ref{alg:navi-feature} demonstrates the procedure to obtain the ideal future volume. 

In H-STGCN, ideal future volume within the prediction window and the corresponding historical average are both taken as input: 
\begin{align}
\begin{split}
\bV_{i, t} = \left[ \nu_{i, t, 0}, \nu_{i, t, 1}, \ldots, \nu_{i, t, F}, \nu^{(h)}_{i, t, 0}, \nu^{(h)}_{i, t + 1, 0}, \ldots, \nu^{(h)}_{i, t + F, 0} \right] ,
\end{split}
\end{align}
where $ i $ is the segment index.

\textbf{Travel Time}.
Travel time $ \tau_{i, t} $ is calculated using the map-matched \cite{lou2009map} GPS data from Amap. In H-STGCN, travel time and its historical average within the prediction window are also both taken as input:
\begin{align}
\begin{split}
\bT_{i, t} = \left[ \tau_{i, t}, \tau^{(h)}_{i, t}, \tau^{(h)}_{i, t + 1}, \ldots, \tau^{(h)}_{i, t + F} \right] ,
\end{split}
\end{align}
where $ i $ is the segment index.

\begin{algorithm}[htbp]
	\caption{Route aggregation algorithm to obtain the ideal future volume}\label{alg:navi-feature}
	\begin{algorithmic}[1]
		\REQUIRE The list of route records from the dataset $ \mathscr{L} $
		\ENSURE Ideal future volume $ \bnu $
		\STATE Initialize $ Z $ as an empty set
		\FORALL {$ r \leftarrow 0, 1, \ldots, N_{\mathscr{L}} - 1 $}
			\FORALL{$ l \leftarrow 0, 1, \ldots, M_{r} - 1 $}
				\STATE {$ s \leftarrow \rho_{r, l} $}
				\quad \COMMENT{$ s $ is the id of a road segment}
				\STATE {$ t \leftarrow \delta_{r, l}$}
				\quad \COMMENT{$ t $ is a time slot}
				\FOR{$ f \leftarrow 0, 1, 2, \cdots, F $}
					\IF{$t \ge \psi_{r}$ }
						\STATE {$ \zeta \leftarrow (s, t, f)$}
						\STATE Add $ \zeta $ to $ Z $
					\ELSE
						\STATE \textbf{break}
					\ENDIF
					\STATE $ t \leftarrow t - \Delta t $ 
					\quad \COMMENT{$ \Delta t $ stands for the length of a single time slot}
				\ENDFOR
			\ENDFOR
		\ENDFOR
		\FORALL {$ s_0 \leftarrow 0, 1, \ldots, n - 1 $}
			\FORALL {$ t_0 \leftarrow 0, 1, \ldots, S_{\text{train}} + S_{\text{test}} - 1 $}
				\FORALL {$ f_0 \leftarrow 0, 1, \ldots, F $}
					\STATE {$ \bnu_{s_0, t_0, f_0} = \text{cardinality}( \{ \zeta | \zeta.s = s_0, \zeta.t = t_0, \zeta.f = f_0, \forall \zeta \in Z \}) $}
				\ENDFOR
			\ENDFOR
		\ENDFOR
		\RETURN $ \bnu $
	\end{algorithmic}
\end{algorithm}

\subsection{Domain Transformer\label{domain_transformer}}
Transformer $ \bLambda $ is proposed to convert ideal future volume into its travel time equivalent. In this manner, any network structure originally designed to process the signal of travel time is equally applicable for volume, easing the process of modality integration. $ \bLambda $ consists of two cascaded layers, the shared $ 1 \times 1 $ convolution and segmentwise $ 1 \times 1 $ convolution, as shown in Figure \ref{fig:model-detail}.

\begin{figure}[h]
	\centering
	\begin{minipage}[t]{1\linewidth}
		\begin{subfigure}[t]{0.99\textwidth}
			\centering
			\includegraphics[width=\linewidth]{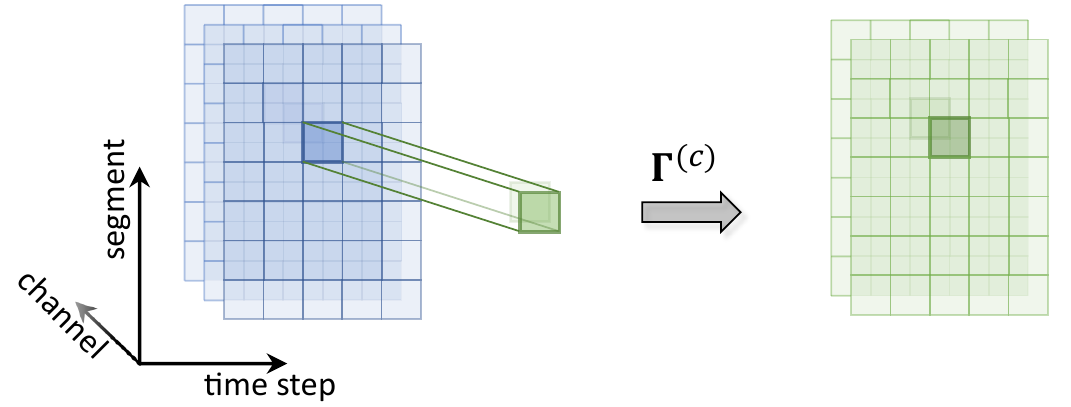} 
			\caption{Shared $ 1\times1 $ convolution}
			\label{fig:shared-conv}
		\end{subfigure}
		\vfill
		\begin{subfigure}[t]{0.99\textwidth}
			\centering
			\includegraphics[width=\linewidth]{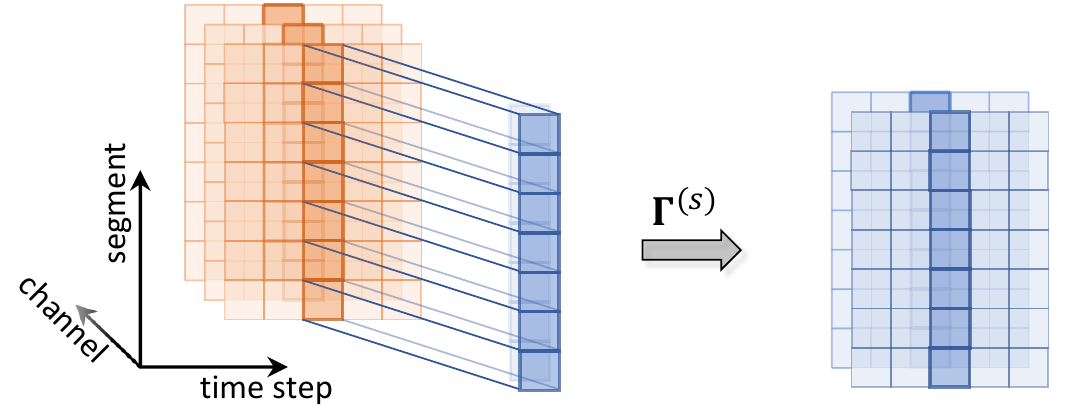} 
			\caption{Segmentwise $ 1\times1 $ convolution}
			\label{fig:segmentwise-conv}
		\end{subfigure}
		\vfill
		\begin{subfigure}[t]{0.99\textwidth}
			\centering
			\includegraphics[width=\linewidth]{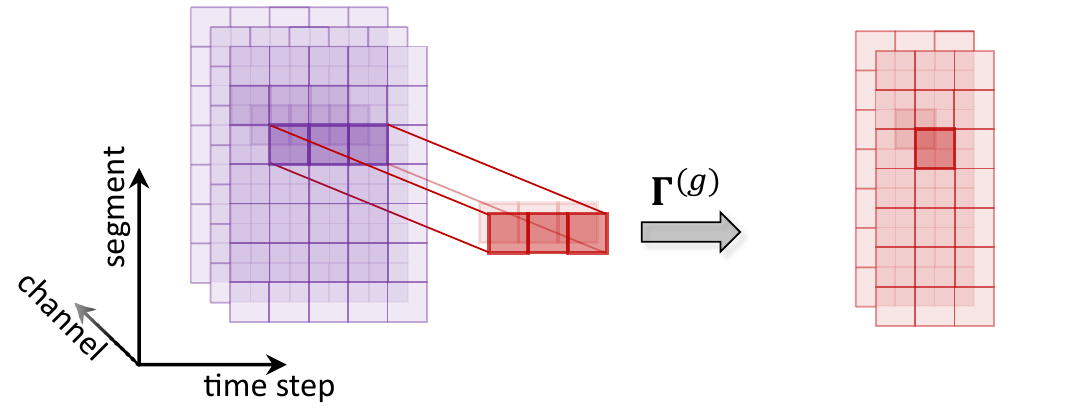} 
			\caption{Temporal gated convolution (with the gate ommited)}
			\label{fig:gated-conv}
		\end{subfigure}
	\end{minipage} 
	\caption{Illustration of convolution operators in H-STGCN. }
	\label{fig:convolution}
\end{figure}

\textbf{Shared $ 1 \times 1 $ Convolution}.
A $ 1 \times 1 $ convolution $ \bGamma^{(c)} $ shared across all segments and time slots is used as the top layer, aiming to capture the universal triangular-shaped mapping. A schematic of the convolution is shown in Figure \ref{fig:shared-conv}. Let $ \bX^{(c)}_{i, t, :}  \in \mathbb{R}^{C^{(c_\text{in})}} $, $ \bY^{(c)}_{i, t, :}  \in \mathbb{R}^{C^{(c_\text{out})}} $ denote the input and output, then this layer works as
\begin{equation}
\bY^{(c)}_{i, t, :} = \bGamma^{(c)}\left(\bX^{(c)}_{i, t, :}\right) = \sigma\left( \bX^{(c)}_{i, t, :} \cdot \bF^{(c)} + \bb^{(c)} \right),
\end{equation}
where $ \bF^{(c)} \in \mathbb{R}^{C^{(c_\text{in})} \times C^{(c_\text{out})}} $ is the weight, $ \bb^{(c)} \in \mathbb{R}^{C^{(c_\text{out})}} $ is the bias, and $ \sigma $ is the Exponential Linear Unit (ELU) \cite{clevert2015fast}.

\textbf{Segmentwise $ 1 \times 1 $ Convolution}.
To ensure sufficient model capacity to extract the segment-level features, a $ 1 \times 1 $ convolution $ \bGamma^{(s)} $ with segment-specific parameters is used as the bottom layer. A schematic of the convolution is shown in Figure \ref{fig:segmentwise-conv}. Let $ \bX^{(s)}_{i, t, :}  \in \mathbb{R}^{C^{(s_\text{in})}} $, $ \bY^{(s)}_{i, t, :} \in \mathbb{R}^{C^{(s_\text{out})}} $ denote the input and output, then this layer works as
\begin{equation}
\bY^{(s)}_{i, t, :} = \bGamma^{(s)}\left(\bX^{(s)}_{i, t, :}\right) = \sigma\left( \bX^{(s)}_{i, t, :} \cdot \bF^{(s)}_{i, :, :} + \bb^{(s)}_{i, :} \right),
\end{equation}
where $ \bF^{(s)} \in \mathbb{R}^{n \times C^{(s_\text{in})} \times C^{(s_\text{out})}} $ is the weight, $ \bb^{(s)} \in \mathbb{R}^{n \times C^{( s_\text{out})}} $ is the bias, and $ \sigma $ is an ELU.

\subsection{Graph Convolution with \\ Compound Adjacency Matrix}
 Graph convolution has been utilized as a key building block in many existing architectures \cite{li2018diffusion,yu2018spatio,fang2019gstnet} to model the non-Euclidean spatial dependency of road traffic. At the core of graph convolution is the weighted adjacency matrix \cite{von2007tutorial}, which as a node proximity measure, determines the spectral modes that are amplified or attenuated by the learnable parameters. Our proposed compound adjacency matrix and the formulation of graph convolution are elaborated as follows.

\textbf{Compound Adjacency Matrix}.
Adjacency matrix in prior works \cite{li2018diffusion,yu2018spatio} assumed a node-proximity with simple exponential distance-decay:
\begin{equation}
w^{(d)}_{ij} = \begin{cases}
\exp \left( - \frac{d_{ij}^2}{\sigma^2} \right) &,
\exp \left( - \frac{d_{ij}^2}{\sigma^2} \right) \ge \epsilon \\
0&, \text{otherwise}
\end{cases} ,
\label{eq:wd}
\end{equation}
where $ d_{ij} $ is the shortest-path distance between segment $ s_i $ and $ s_j $, $ \sigma $ denotes the spatial attenuation length, and $ \epsilon $ is a cutoff controlling the matrix sparsity. We call $ \bW^{(d)} $ the Dijkstra matrix in the following. This pure spatial closeness fails to reflect the actual traffic proximity in many scenarios. To be specific, the effect of an occurrence of congestion on traffic diversion depends on several attributes of the adjacent road segment, including the functional class, pavement condition, etc. Thus, congestion propagation is often not spatially uniform, contradicting the aforementioned assumption. To overcome this problem, we propose the compound adjacency matrix $ \bW^{(c)} $ as follows:
\begin{align}
	\begin{split}
		w^{(c)}_{ij} &= \sigma_{ij} \cdot w^{(d)}_{ij}, 1 \le i \le n,1 \le j \le n ,\\
		\sigma_{ij} &= \sum_{t \in [0, S_{\text{train}})} \left(\tau_{i, t} - \bar{\tau}_i\right)_+\left(\tau_{j, t} - \bar{\tau}_j \right)_+ ,
	\end{split}
	\label{eq:cam}
\end{align}
where $ (\cdot)_+ = \max\left\{ 0, \cdot \right\} $, $ \bar{\tau}_i = \sum_{t \in [0, S_{\text{train}})} \tau_{i,t} / S_{\text{train}} $. Term $ \sigma_{ij} $ is the equivalent of the travel time correlation between segment $ i $ and $ j $ subtracting the $ (\cdot)_+ $ operation. This operation is added to remove the ``correlation floor'' derived from the common free-flow periods. In this paper, $ \Sigma $ is referred to by covariance matrix for convenience.

As shown in Eqn. \ref{eq:cam}, the compound adjacency matrix is the Hadamard product of the covariance matrix and the Dijkstra adjacency matrix. The incorporation of the covariance term is inspired by the connection between graph convolution and the standard convolutional neural network (CNN) widely utilized in computer vision tasks. As pointed out in \citep{bruna2014spectral}, when applied to natural images, a graph convolution using the covariance of pixel intensity as proximity measure recovers a standard CNN without any prior knowledge. The covariance term $ \Sigma $ is therefore analogously presumed to offer a more intrinsic measure for traffic proximity. Meanwhile, the Dijkstra matrix is retained to eliminate the unphysical long-range correlations in $ \Sigma $, such as those induced by citywide daily rush-hour congestion.

\textbf{Graph Convolution}.
The regional road network is considered as an undirected graph, with each node representing a particular road segment. As in the prior work \citep{yu2018spatio}, a shared graph convolutional network (GCN) is applied on each individual time slice to extract common spatial patterns, and we implement GCN with the spectral formulation \cite{defferrard2016convolutional}. Specifically, we have the normalized graph Laplacian $ \bL $ and scaled graph Laplacian $ \tilde{\bL} $ as 
\begin{align}
	\bL &= \bI_n - \bD^{-\frac{1}{2}} \bW^{(c)} \bD^{-\frac{1}{2}} , \\
	\tilde{\bL} &= 2 \bL / \lambda_{\max} - \bI_n ,
\end{align}
where $ \bI_n $ is the identity matrix, $\bW^{(c)}$ is the compound adjacency matrix, $ \bD $ is the diagonal degree matrix of $\bW^{(c)}$ with $ D_{ii} = \sum_{j=1}^n w^{(c)}_{ij} $, and $ \lambda_{\max} $ is the greatest eigenvalue of $ \bL $. The GCN $ \bTheta $ is parametrized with Chebyshev polynomials of the scaled graph Laplacian $ \tilde{\bL} $. Let $ \bX_{:, t, :}^{(\bTheta)}  \in \mathbb{R}^{n\times C^{(\bTheta_\text{in})}} $, $ \bY_{:, t, :}^{(\bTheta)}  \in \mathbb{R}^{n\times C^{(\bTheta_\text{out})}} $ denote the input and output, then GCN $ \bTheta $ works as:
\begin{align}
\begin{split}
	\bY_{:, t, j}^{(\bTheta)} = \sigma \left( \sum_{m=1}^{C^{(\bTheta_\text{in})}} \sum_{k=0}^{K-1} \bTheta_{k, m, j} T_k(\tilde{\bL})  \bX^{(\bTheta)}_{:, t, m} + \bb^{(\bTheta)}_j \right) \in \mathbb{R}^{n}, \\ 
	\forall j = 1, 2, \ldots, C^{(\bTheta_\text{out})}
\end{split}
\end{align}
where $ T_k(\tilde{\bL}) $ is the $ k $-th order Chebyshev polynomial, $ K $ is the kernel size, $  \bTheta \in \mathbb{R}^{K \times C^{(\bTheta_\text{in})} \times C^{(\bTheta_\text{out})}} $ denotes the parameter tensor, $ \bb_j $ is the bias, and $ \sigma $ is an ELU.

\subsection{Temporal Gated Convolution}
To extract common temporal features, we take advantage of the temporal gated convolution $ \bGamma^{(g)} $ proposed in \citep{yu2018spatio}. As illustrated in Figure \ref{fig:gated-conv}, a shared gated 1D convolution is applied on each road segment along the temporal dimension. The 1D convolution maps the input $ \bX^{(g)} \in \mathbb{R}^{n\times P \times C^{(g_\text{in})}} $ to a tensor:
\begin{align}
 \left[ \bA \text{ }\bB\right] \in \mathbb{R}^{n\times(P - K_t + 1) \times (2 C^{(g_\text{out})})} = \bF^{(g)} * \bX^{(g)} + \bb^{(g)} ,
\end{align}
where $ * $ is the 1D-convolution operator, $ \bF^{(g)} \in \mathbb{R}^{K_t \times C^{(g_\text{in})} \times 2 C^{(g_\text{out})}} $ is the convolution kernel, $ K_t $ is the kernel size, $ P $ is the length of input temporal sequence, $ \bb^{(g)} $ is the bias, and $ \bA $ and $ \bB $ are of equal size with $ C^{(g_\text{out})} $ channel. A gated linear unit (GLU) with $ \bA $ and $ \bB $ as inputs further adds non-linearity to obtain this layer's output: $ \bGamma^{(g)} \left( \bX^{(g)} \right) = \bA \odot \sigma(\bB) \in \mathbb{R}^{n\times(P - K_t + 1) \times C^{(g_\text{out})}} $. ``$ \odot $'' stands for the operator of element-wise multiplication.

\subsection{Connection to STGCN}
Proposed by \citet{yu2018spatio}, the Spatio-Temporal Graph Convolutional Network (STGCN) stacks the spatial graph convolutional layer and temporal gated convolutional layer multiple times in an alternating fashion to jointly capture spatio-temporal dependency. When dropping the volume-feature processing branch (the hybrid branch) and the covariance term in the adjacency matrix, our proposed model reduces to a STGCN model with a single ST-Conv block.

\subsection{Model Training\label{modeltraining}}
\textbf{Data Augmentation}.
Since traffic volume is discrete in nature, in situations of low traffic, even a small fluctuation in the volume channel would considerably affect the model output, making it hard to generalize. To solve this problem, Gaussian noise is added \citep{goodfellow2016deep} on all volume channels with values below a threshold $\epsilon_n $. Experiment results show that this data augmentation approach significantly mitigates the overfitting.

\textbf{Optimization}.
For the multistep traffic forecasting task in this paper, we use the L1 loss function:
\begin{align}
	\begin{split}
		\mathcal{L} = \dfrac{1}{n \times S_{\text{train}} \times F}\sum_{\substack{i \in [0, n) \\ t \in [0, S_{\text{train}}) \\ f \in (0, F]}}{|\hat{\tau}_{i, t + f} - \tau_{i, t + f}|} ,
	\end{split}
\end{align}
where $ \hat{\tau}_{i, t + f} $ is the model output and $ \tau_{i, t + f} $ is the ground truth.

\section{EXPERIMENTS\label{experiments}}
In this section, we first describe the datasets, compared methods, implementation details, and evaluation metrics. Then we show the effectiveness of the compound adjacency matrix, future-volume feature, and domain transformer. At last, we discuss the model scalability.

\subsection{Datasets\label{dataset}}
\begin{figure}[h]
	\centering
	\begin{minipage}[t]{1\linewidth}
		\begin{subfigure}[t]{0.48\textwidth}
			\centering
			\includegraphics[width=\linewidth]{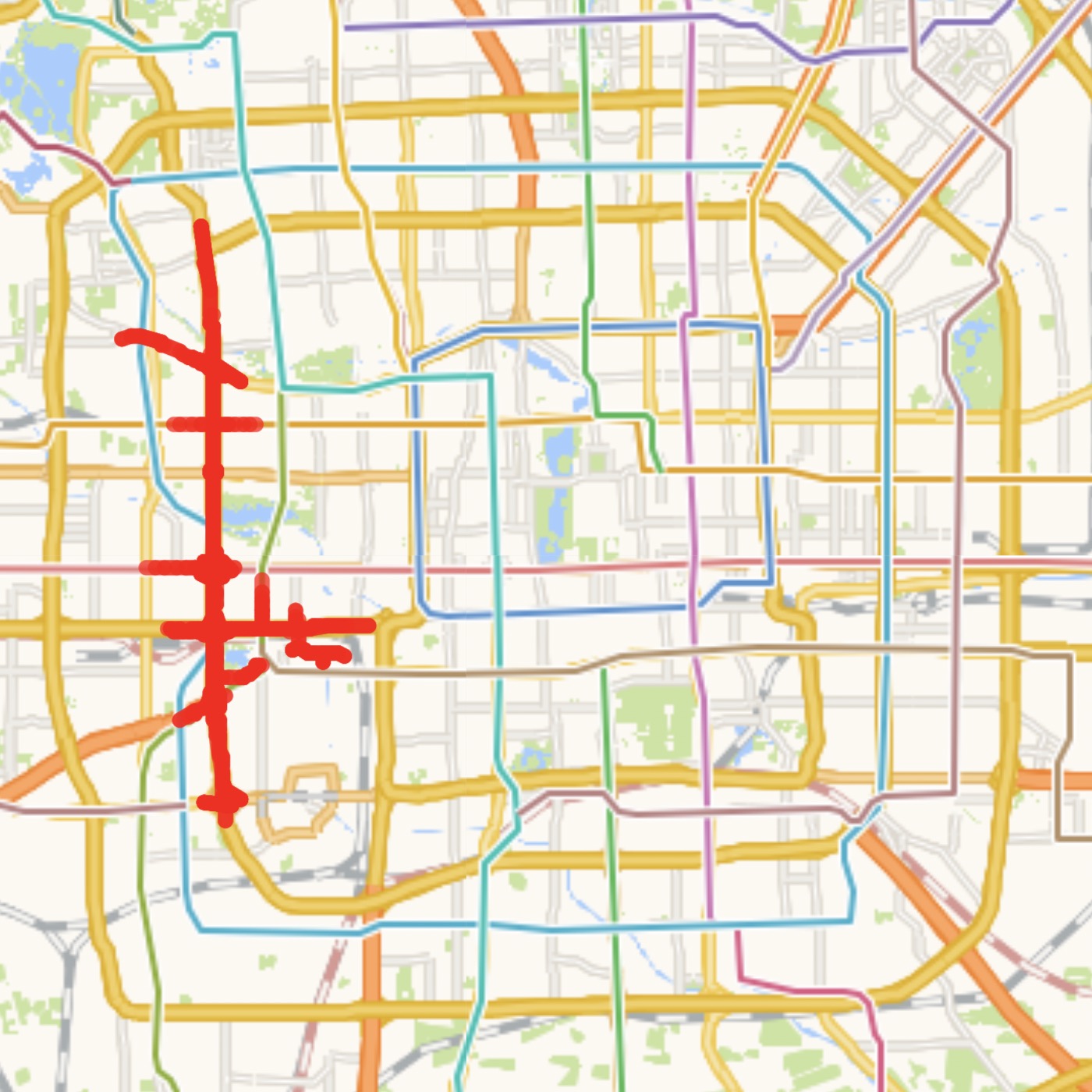} 
			 \caption{Road segments of W3-715} 
		\end{subfigure}
		\hfill
		\begin{subfigure}[t]{0.48\textwidth}
			\centering
			\includegraphics[width=\linewidth]{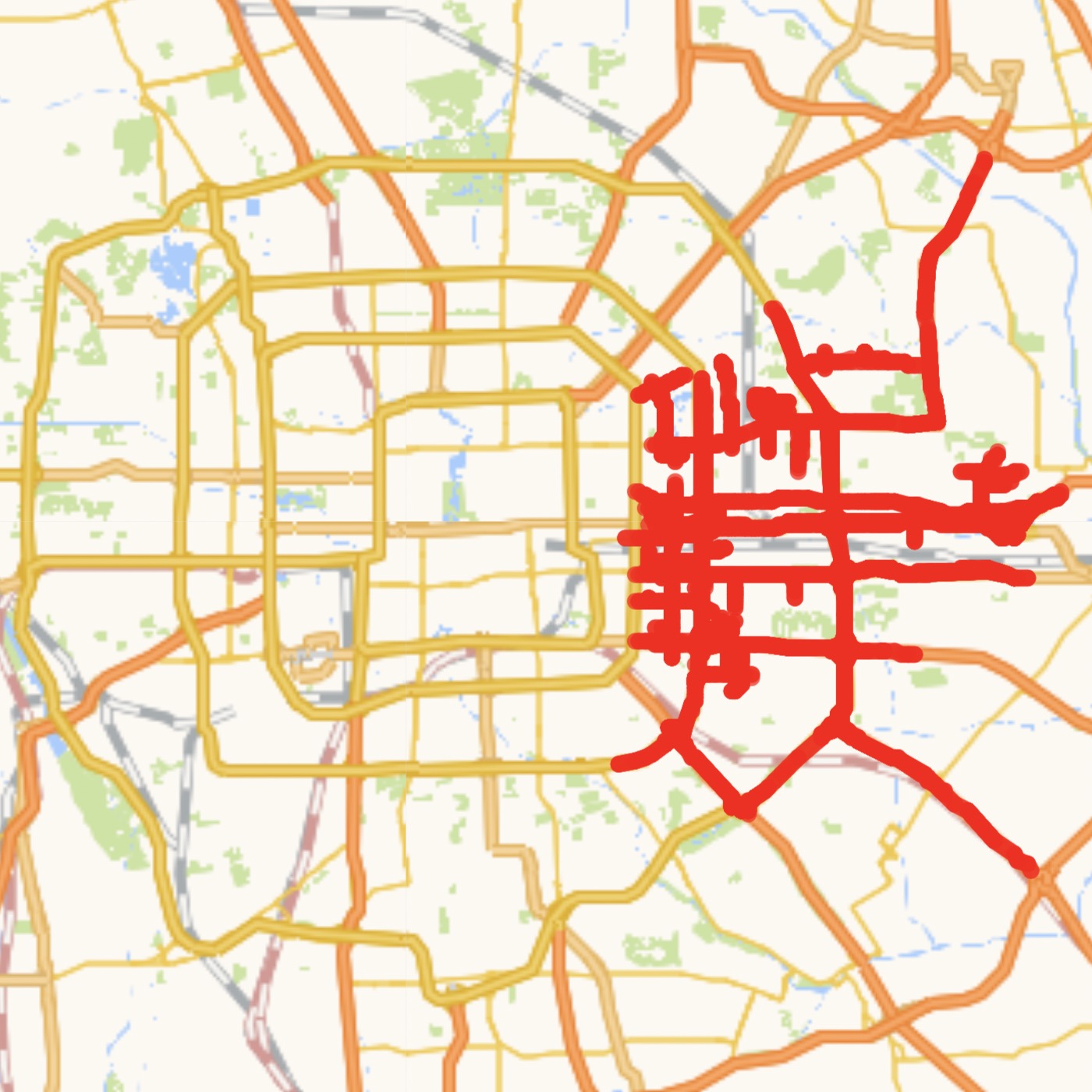} 
			\caption{Road segments of E5-2907} 
		\end{subfigure}
	\end{minipage} 
	\caption{Spatial distribution of the regional road networks.}
	\label{fig:road-segments}
\end{figure}

Using anonymous user data from Amap, we conduct experiments on two regional networks in the Beijing area as shown in Figure \ref{fig:road-segments}: one is around the West 3rd Ring Road with 715 segments, and the other around the East 5th Ring Road with 2907 segments. The respective datasets are denoted by W3-715 and E5-2907. Table \ref{tab:road-stat} depicts the statistics of road segments in the two networks.

Each dataset contains traffic condition and navigation records from 06:00 to 22:00, and the time span is from December 24, 2018 to April 21, 2019 with holidays removed (ten weeks in total). The previous eight weeks are used as training data, and the remaining two weeks as testing data.

\begin{table}[htbp]
	\centering
	\caption{Statistics of road segments in the network, including the total road segment number, the average length (meter) of road segments, and the average traffic volume (veh/min) of road segments corresponding to each road class.}
	\begin{tabular}{ccrrr}
		\toprule
		Dataset           &    Road class    & Num. & Avg. len. (meter) & Avg. vol.  \\
		\midrule
		\multirow{5}[2]{*}{W3-715}  
		&     Freeway      		 &   7    &       132        & 7.0 \\
		& Highway	&   34   &      176        & 3.2 \\
		&    Expressway		  &  186   &       163        & 11.8 \\
		&    Major Road    &  488  &        80        & 2.4 \\
		&      Total       &  715   &       107        & 4.9 \\ 
		\midrule
		\multirow{5}[2]{*}{E5-2907} 
		&     Freeway      &   135   &       334        & 9.9 \\
		& Highway &  163   &       178       			& 2.6 \\
		&    Expressway    &  427   &       348        &12.4 \\
		&    Major Road    &  2182  &        97       & 2.3 \\
		&      Total       &  2907  &       150        & 4.1 \\ 
		\bottomrule& 
	\end{tabular}%
	\label{tab:road-stat}%
\end{table}%

\begin{table}[htbp]
	\centering
	\caption{Statistics of high volume road segments, including the number of road segments, the percentage of congested periods (C), the percentage of non-recurring congested periods (NRC), and the average traffic volume (veh/min).}
	\begin{tabular}{ccrrrr}
		\toprule
		Dataset           &    Road class    & Num. & Pct. (C) & Pct. (NRC)  & Avg. vol.  \\ 
		\midrule
		\multirow{4}[2]{*}{\shortstack{W3-715}}  
		& Freeway      	  &   0    & / &    /    & / \\
		& Highway	      &  0 & / &       /   & / \\
		& Expressway	&  138   & 19.6\% & 6.5\%       & 14.0\\
		&    Major Road    &  1  & 22.9\% & 5.4\%       & 10.2 \\ 
		\midrule
		\multirow{4}[2]{*}{\shortstack{E5-2907}} 
		&     Freeway      &   70   & 7.8\% & 3.4\%       & 12.1 \\
		& Highway &  5   & 11.5\% & 8.5\%    				& 15.7 \\
		&    Expressway   &  235   & 15.7\% & 7.3\%       & 18.0 \\
		&    Major Road   &  1  & 39.3\% & 21.3\%     & 10.2 \\
		\bottomrule
	\end{tabular}%
	\label{tab:hv-road-stat}%
\end{table}%

\subsection{Compared Methods}

We compare our proposed architecture with the following two methodological categories:

\textbf{Benchmark Models}.
\begin{itemize}
	\item Historical Average (HA): Historical average predicts travel time with mean value over time slots at the same previous relative position.
	\item Linear Regression (LR): Linear regression is a basic regression model.
	\item Gradient Boosting Regression Tree (GBRT): GBRT is a widely-used boosting model. We set the number of trees at 50, with a maximum depth of 6.
	\item Multi-Layer Perceptron (MLP): MLP is a fully connected multi-layer neural network. We use three layers, and the hidden unit of each layer is 64.
	\item Sequence-to-Sequence (Seq2Seq): Seq2Seq models use the encoder-decoder architecture and have been widely applied to language modeling and time-series forecasting. We use two layers, and the hidden unit of each layer is 200.
	\item STGCN: Original STGCN used multiple ST-Conv blocks to boost the performance. On our dataset, however, one block is found sufficient to achieve a similar level of accuracy. We thus use STCGN with a single ST-Conv block  as the baseline.
\end{itemize}

\textbf{Variant Models for Ablation Study}.
\begin{itemize}
	\item STGCN (Im): Improved STGCN uses a compound adjacency matrix as opposed to the Dijkstra matrix.
	\item H-STGCN (1): H-STGCN (1) uses an input volume tensor $ \bV $ with all elements set to one (1).
\end{itemize}

\subsection{Implementation Details}
In all models, we use data from the previous six time slots (30 min) as input, and predict travel time for the next hour. The shared and segmentwise $ 1 \times 1 $ convolutions in domain transformer both have 16 filters. The graph convolution has 64 filters. The temporal gated convolutional layers $ \bGamma^{(g)}_1 $, $ \bGamma^{(g)}_2 $, $ \bGamma^{(g)}_3 $, $ \bGamma^{(g)}_4 $ have 64, 128, 64, and 64 filters. The last fully connected layer outputs 12 values, corresponding to the forecasting period. We set $ \sigma^2 = \text{3 km}^2 $, $ \epsilon = 0$ (no spatial cutoff) in Equation (\ref{eq:wd}), and the threshold of noise injection $ \epsilon_n = 3 $. We use Adam optimizer \citep{kingma2014adam} with initial learning rate 0.001 and decay rate 0.98. We implement the traditional benchmark models with scikit-learn, and the neural network models on TensorFlow. The training and inference of neural networks are conducted on 4 NVIDIA GPUs with 16 GB memory.

\subsection{Evaluation Metrics}
To better verify the effectiveness of H-STGCN, we select two additional subtestsets based on the following considerations. First, to showcase the extra predictive power brought by ideal future volume, we consider only segments with average historical traffic volume above 10 veh/min (high-volume segments). Secondly, we focus only on congested time periods, the non-trivial part of traffic forecasting. The congestion speed threshold is set according to road class: 30 km/h for freeway, 20 km/h for highway and expressway, and 12 km/h for major road. We further define non-recurring congestion as the one with travel speed constantly below half of its historical average. Thirdly, to examine forecasting performance over the full lifecycle of congestion, we extend a (non-recurring) congested period by an hour in each direction to include both the formation and dissipation stages of congestion. To summarize, we have three types of test sets in the experiment:
\begin{itemize}
	\item Full test set as described in Section \ref{dataset}.
	\item Test set comprising data from high-volume segments in the congested periods, denoted by suffix (C).
	\item Test set comprising data from high-volume segments in the non-recurring congested periods, denoted by suffix (NRC).
\end{itemize}
Table \ref{tab:hv-road-stat} shows statistics of the last two test sets. We use the mean absolute error (MAE), mean absolute percentage error (MAPE), and root mean square error (RMSE) as the evaluation metrics.

\begin{table*}[t!]
	\centering
	\caption{Comparison with baselines on full test set, test set (C), and test set (NRC). Evaluation metrics include MAE (s/m), MAPE (\%), and RMSE (s/m).}
	\begin{tabular}{c|c|ccc|ccc|ccc}
		\toprule
		Dataset & Model & MAE  & MAPE  & RMSE  & MAE  & MAPE  & RMSE & MAE  & MAPE  & RMSE \\
		\midrule
		& &\multicolumn{3}{c|}{Test set (Full)} &\multicolumn{3}{c|}{Test set (C)} &\multicolumn{3}{c}{Test set (NRC)} \\
		\midrule
		\multirow{10}[2]{*}{\shortstack{W3-715}} 
		&HA    & 0.03886 & 20.73 & 0.09285 & 0.07040 & 34.36 & 0.10479 & 0.10303 & 39.39 & 0.16486 \\
		&LR    & 0.03334 & 16.58 & 0.08467 & 0.06469 & 33.52 & 	0.10582 & 0.09080 & 39.57 & 	0.14768 \\
		&GBRT  & 0.03264 & 16.10 & 0.08409 & 0.06236 & 32.08 & 0.10479 & 0.09085 & 39.35 & 0.14945 \\
		&MLP   & 0.03272 & 16.57 & 0.08269 & 0.06096 & 31.84 & 0.10190 & 0.08733 & 38.71 & 0.14427 \\
		&Seq2Seq & 0.03231 & 15.81 & 0.08252 & 0.06033 & 28.79 & 0.10174 & 0.08599 & 34.04 & 0.14467 \\
		&STGCN & 0.03219 & 16.01 & 0.08182  & 0.05975 & 30.48 & 0.09901 & 0.08599 & 38.72 & 0.14004 \\
		\cmidrule{2-11}
		&STGCN (Im) & 0.03200 & 15.83 & 0.08196 & 0.05965 & 29.96 & 0.09995 & 0.08539 & 36.71 & 0.14197 \\
		&H-STGCN (1) & 0.03138 & 15.52 & 0.08099 & 0.05804 & 29.14 & 0.09806 & 0.08373 & 34.71 & 0.14012 \\
		\cmidrule{2-11}
		&H-STGCN & \textbf{0.03114} & \textbf{15.36} & \textbf{0.08045} & \textbf{0.05711} & \textbf{28.34} & \textbf{0.09644} & \textbf{0.08124} & \textbf{33.22} & \textbf{0.13711} \\
		\cmidrule{1-11}
		\multirow{10}[2]{*}{\shortstack{E5-2907}} 
		&HA    & 0.04615 & 21.22 & 0.11405 & 0.09786 & 44.95 &0.16729 & 0.13161 & 46.96 & 0.21769 \\
		&LR    & 0.04096 & 17.03 & 0.10732 & 0.08229 & 41.69 & 	0.14270 & 0.10747 & 47.01 & 	0.18192 \\
		&GBRT  & 0.04032 & 16.61 & 0.10680 & 0.07997 & 39.51 & 0.14465 & 0.10657 & 44.68 & 0.18593 \\
		&MLP   & 0.04031 & 17.16 & 0.10547 & 0.08025 & 41.26 & 0.14229 & 0.10580 & 45.84 & 0.18236 \\
		&Seq2Seq & 0.04087 & 17.52 & 0.10631 & 0.08413 & 41.72 & 0.14703 & 0.10981 & 44.81 & 0.18722 \\
		&STGCN & 0.03984 & 16.95 & 0.10296 & 0.07561 & 38.13 & 0.13677 & 0.09966 & 43.28 & 0.17563 \\
		\cmidrule{2-11}
		&STGCN (Im) & 0.03957 & 16.85 & 0.10221 & 0.07498 & 37.80 & 0.13579 & 0.09843 & 42.74 & 0.17399 \\
		&H-STGCN (1) & 0.03870 & 16.31 & 0.10095 & 0.07380 & 37.07 & 0.13455 & 0.09750 & 42.32 & 0.17257 \\
		\cmidrule{2-11}
		&H-STGCN & \textbf{0.03861} & \textbf{16.28} & \textbf{0.10067} & \textbf{0.07254} & \textbf{36.31} & \textbf{0.13308} & \textbf{0.09528} & \textbf{40.82} & \textbf{0.17030} \\
		\bottomrule
	\end{tabular}%
	\label{tab:performance-comparison}
\end{table*}

\subsection{Performance Comparison}
Table \ref{tab:performance-comparison} outlines the performance of our model as compared to the competing methods. H-STGCN significantly outperforms the various benchmarks in all metrics, especially for the prediction of non-recurring congestion. In this section, we study the effectiveness of each proposed module in H-STGCN.

\begin{figure}[h]
	\centering
	\begin{minipage}[t]{1\linewidth}
		\begin{subfigure}[t]{0.32\textwidth}
			\centering
			\includegraphics[width=\linewidth]{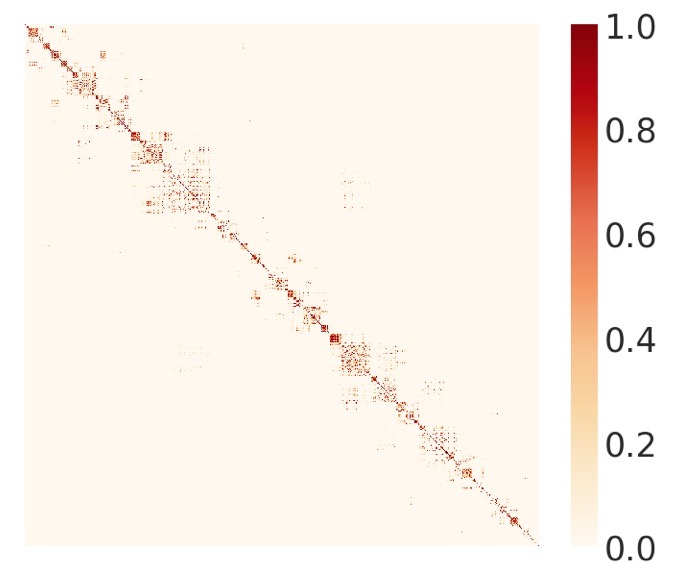} 
			\caption{}
		\end{subfigure}
		\hfill
		\begin{subfigure}[t]{0.32\textwidth}
			\centering
			\includegraphics[width=\linewidth]{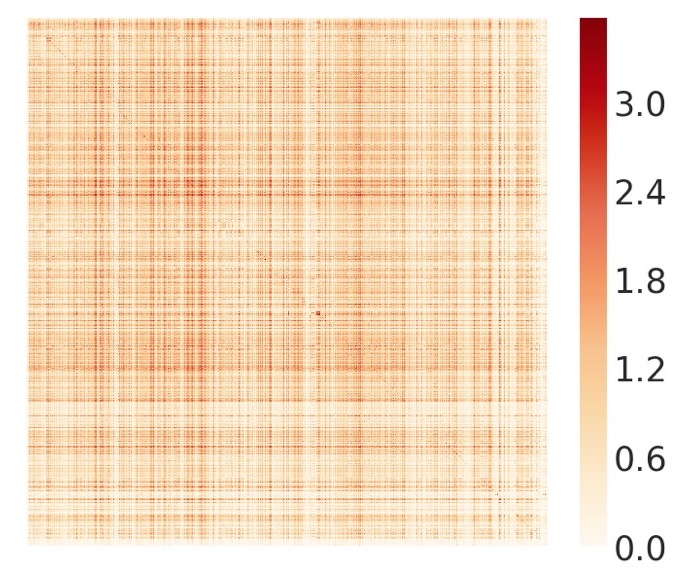} 
			\caption{}
		\end{subfigure}
		\hfill
		\begin{subfigure}[t]{0.32\textwidth}
			\centering
			\includegraphics[width=\linewidth]{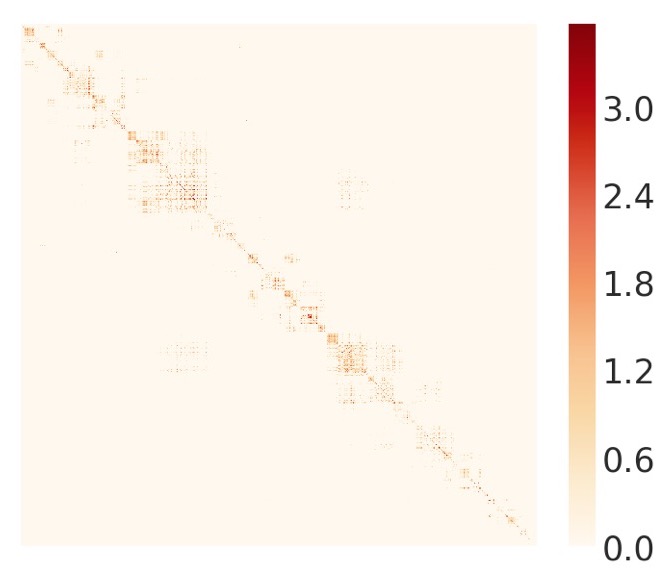} 
			\caption{}
		\end{subfigure}
	\end{minipage} 
	\caption{Weighted adjacency matrices in E5-2907. The color represents the normalized value of $ \lg \left(w_{ij} + 1\right) $. (a) Dijkstra adjacency matrix. (b) Covariance matrix. (c) Compound adjacency matrix. }
	\label{fig:weight-matrix}
\end{figure}

\textbf{Compound Adjacency Matrix}.
We compare the performance of STGCN and STGCN (Im). As shown in Table \ref{tab:performance-comparison}, STGCN (Im) achieves a lower MAE and MAPE on W3-715, and a lower MAE, MAPE, and RMSE on E5-2907, validating the effectiveness of the compound adjacency matrix. Figure \ref{fig:weight-matrix} shows an example from E5-2907, which illustrates the connections among different adjacency matrices as described in Section \ref{domain_transformer}.

\begin{figure}[h]
	\centering
	\begin{minipage}[t]{1\linewidth}
		\begin{subfigure}[t]{0.5\textwidth}
			\centering
			\includegraphics[width=\linewidth]{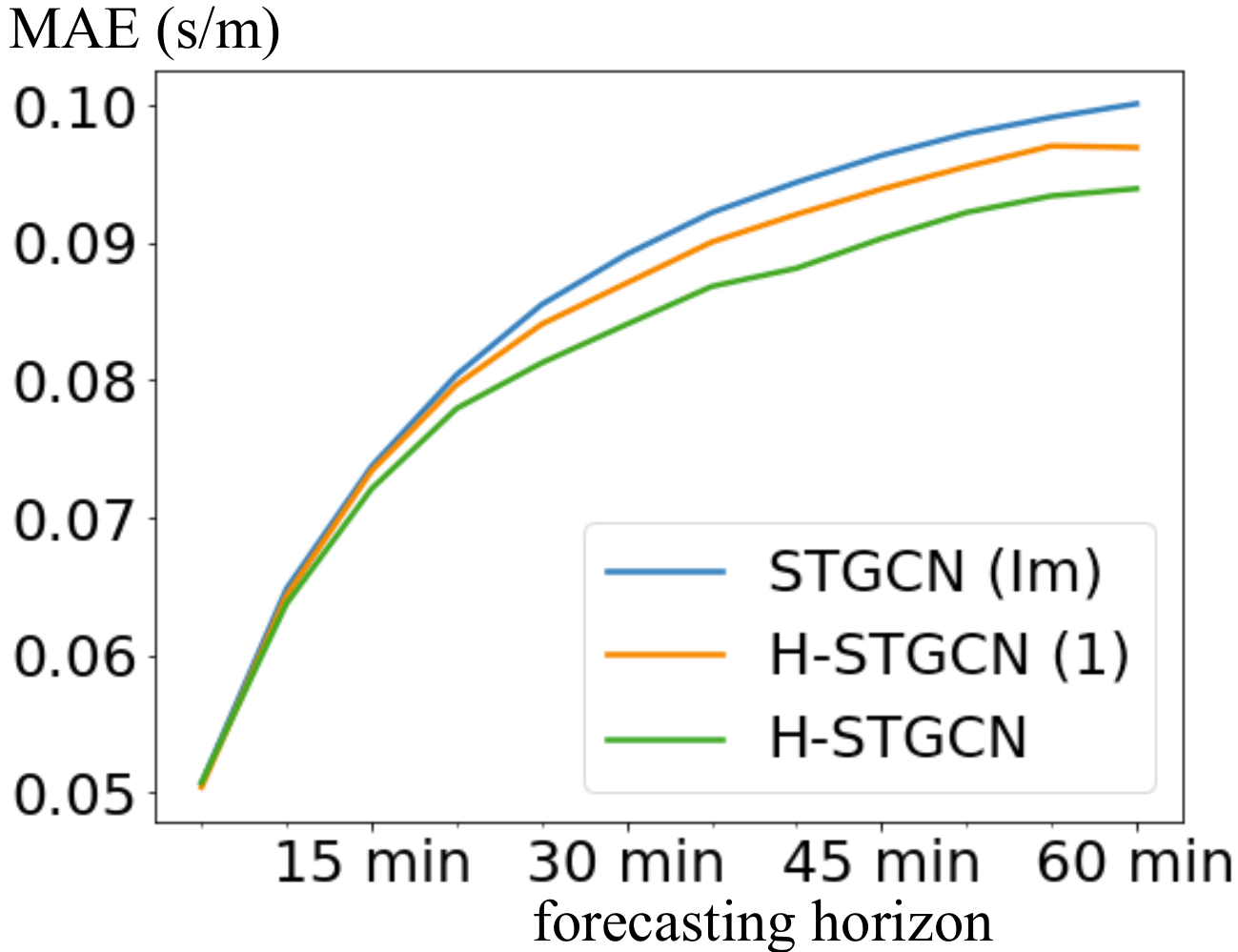} 
			\caption{Dataset W3-715}
		\end{subfigure}
		\hfill
		\begin{subfigure}[t]{0.5\textwidth}
			\centering
			\includegraphics[width=\linewidth]{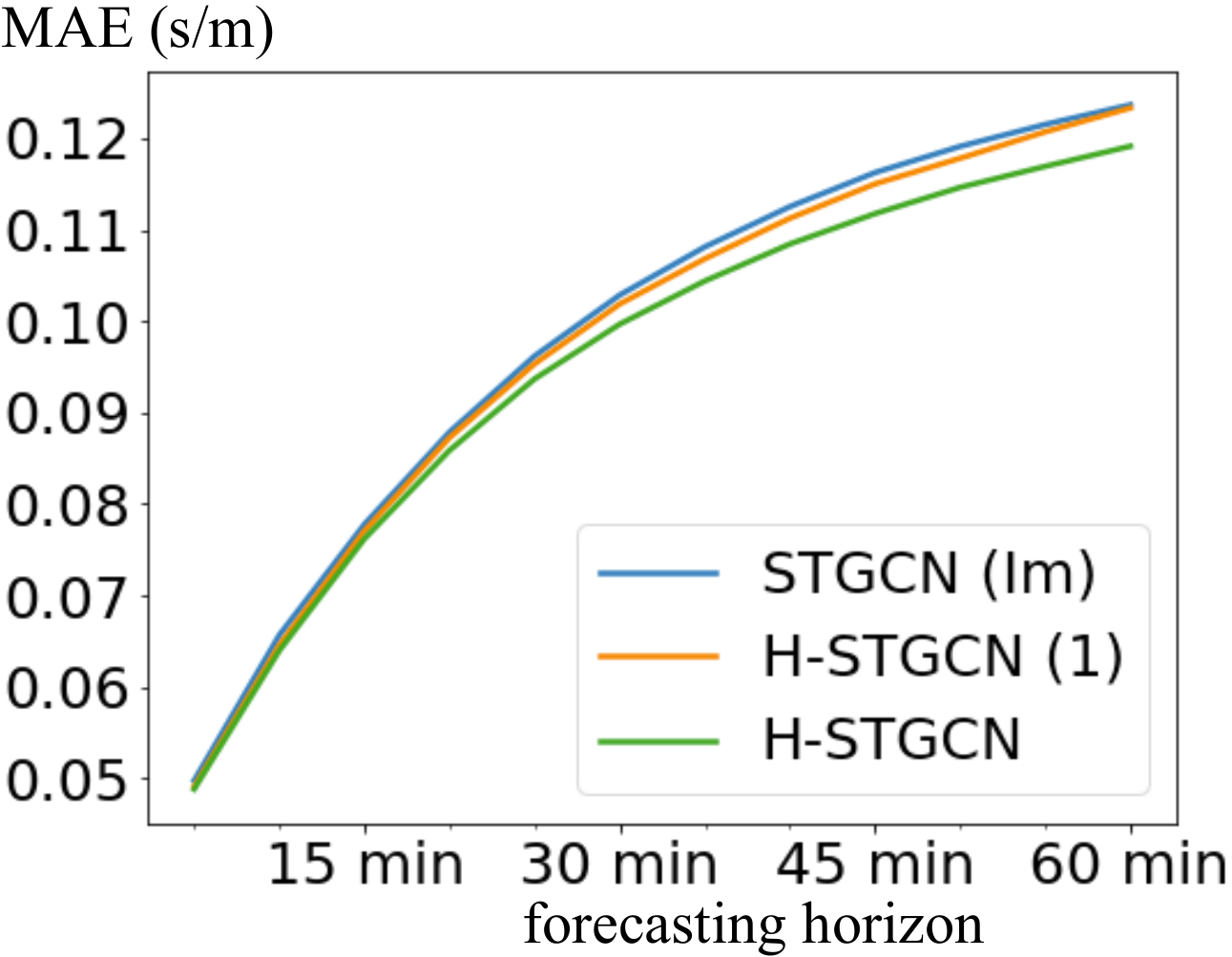} 
			\caption{Dateset E5-2907}
		\end{subfigure}
	\end{minipage} 
	\caption{Comparison over the forecasting horizon on test set (NRC).}
\label{fig:step-mae}
\end{figure}

\textbf{Future-Volume Feature and Domain Transformer}.
First, as shown in Table \ref{tab:performance-comparison}, H-STGCN delivers consistently superior performance compared to STGCN (Im), demonstrating the remarkable advantage brought by the utilization of future-volume data. Secondly, owing to the segment-wise structure of domain transformer, H-STGCN gains an edge over STGCN (Im) in terms of representation power. To eliminate this influence factor and assess the importance of the future-volume feature, we further compare H-STGCN to H-STGCN (1). As indicated by the results on test set (C) and test set (NRC), the volume feature substantially enhances model performance on congestion forecasting. Lastly, Figure \ref{fig:step-mae} shows that, as the forecasting horizon lengthens, the volume feature becomes the most dominant contributor to error reduction.

\begin{figure}[t!]
	\centering
	\includegraphics[width=0.95\linewidth]{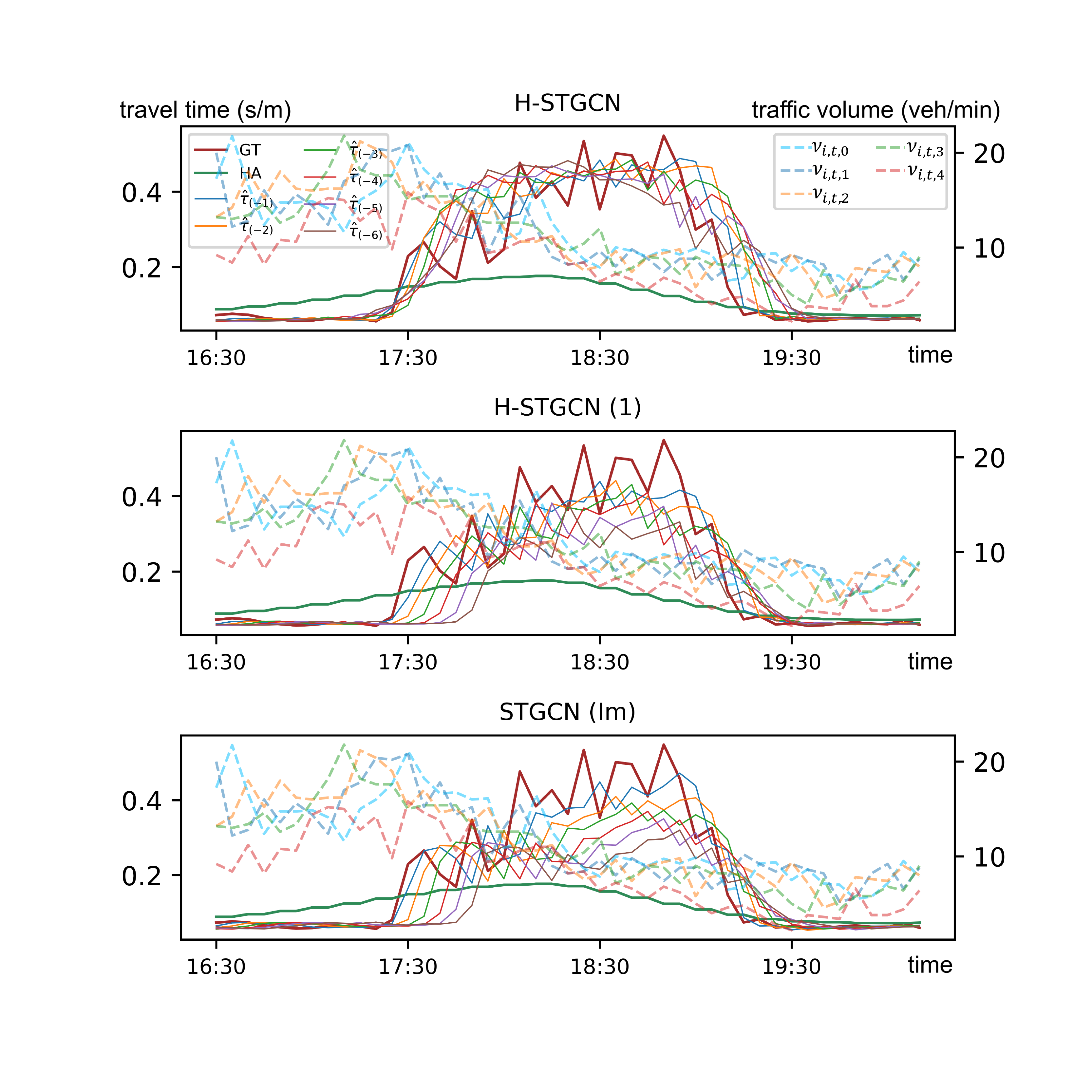}
	\caption{ Example of travel time prediction on non-recurring congestion. Case studied is from a freeway segment on April 16, 2018. GT denotes the ground truth, HA denotes the historical average, $ \hat{\tau}_{(-f)} $ is the $ f $-step ahead prediction, and $ \nu_{i, t, f} $ is the ideal future traffic volume corresponding to the time slot $ f $-step later.
	}
	\label{fig:prediction-case}
\end{figure}

To explain the intuition behind H-STGCN, we show an example regarding the prediction of non-recurring congestion, as depicted in Figure \ref{fig:prediction-case}. During the congestion formation stage between 17:30 and 18:00, the multistep-ahead travel-time prediction from H-STGCN (1) shows a notable time lag compared to the ground truth. In contrast, H-STGCN, when fed with the ideal-future-volume data, is able to accurately forecast the congestion even 30 minutes in advance. We understand this observation as follows. The curve of $ \nu_{i, t, 3} $, which represents an approximation of the traffic volume 15 minutes later, rapidly increases at around 17:15. Further, given the fact that a navigation engine is only aware of the trips that have started already, the actual future traffic volume would be even greater than the ideal one. Therefore, the rise of $ \nu_{i, t, 3} $ is a prominent indicator of strong upcoming traffic flux, which in turn enables H-STGCN to foresee the future congestion even without a historical reference.

\subsection{Model Scalability}
The model inference time for W3-715 and E5-2097 is less than 100 ms. To balance the inference efficiency and forecasting performance for real-world application, we partition a city-wide road network into sub-networks with at most a few thousand segments, by minimizing the number of congested boundary links \cite{karypis1998fast}. Then a separate model is trained and deployed for each sub-network.

\section{RELATED WORK\label{related work}}
Traffic prediction has been studied for decades, and existing methods chiefly fall into two categories: the theory-driven approach and the data-driven approach. In the former category \cite{ben1998dynamit,burghout2004hybrid,vlahogianni2015computational}, a simulation system is built according to the theory of traffic dynamics and is composed of several interacting modules such as a routing model, driving behavior model, and queueing model. Given all origin-destination pairs, a simulator is able to forecast future traffic. For the data-driven approach, shallow machine learning models, including Bayesian network \cite{pascale2011adaptive}, support vector regression, random forest, gradient boosting regression tree etc., were thoroughly investigated. However, due to limited representation power, such elementary models cannot yield the prospective outcomes.

Recently, a host of deep-learning-based approaches have been attempted and achieved considerable improvements over traditional benchmarks. To capture the spatial dependency, graph convolution structures have been subject to experimentation and achieved notable improvements \cite{yu2018spatio,li2018diffusion}. To further extract global spatial correlations, \citet{fang2019gstnet} proposed the use of a non-local correlated mechanism. To model the nonlinear temporal dependency, \citet{li2018diffusion} applied the encoder-decoder architecture. \citet{yu2018spatio} considered time series of traffic speed as a one-dimensional image and instead adopted the convolutional network. 

Amongst the various traffic scenarios, non-recurring congestion is particularly difficult to predict due to the lack of contextual information. To tackle this issue, several studies suggested the use of weather status, tweets, road structure features, points of interest, or crowd map queries as auxiliary information and achieved decent performance \cite{dailey2006use,liao2018deep,he2013improving,koesdwiady2016improving,zheng2019deepstd}. Nonetheless, the spatial resolution of forecasting is insufficient for critical real world application.

\section{CONCLUSION\label{conclusion}}

In this paper, we propose a novel deep architecture for travel time forecasting, the Hybrid Spatio-Temporal Graph Convolutional Network (H-STGCN), which features the utilization of intended-traffic-volume data. We design the domain transformer to couple this heterogeneous modality of traffic volume. We propose a compound adjacency matrix to capture the innate nature of traffic proximity. Experiments carried out on real-world datasets show that H-STGCN achieves remarkable improvement over the benchmark methods, especially for the prediction of non-recurring congestion. Finally, this architecture exemplifies a novel formalism to embed the knowledge of physics in a data-driven model, which can be readily applied to general spatio-temporal forecasting tasks.

\bibliographystyle{ACM-Reference-Format}
\bibliography{references}


\begin{thebibliography}{26}


\ifx \showCODEN    \undefined \def \showCODEN     #1{\unskip}     \fi
\ifx \showDOI      \undefined \def \showDOI       #1{#1}\fi
\ifx \showISBNx    \undefined \def \showISBNx     #1{\unskip}     \fi
\ifx \showISBNxiii \undefined \def \showISBNxiii  #1{\unskip}     \fi
\ifx \showISSN     \undefined \def \showISSN      #1{\unskip}     \fi
\ifx \showLCCN     \undefined \def \showLCCN      #1{\unskip}     \fi
\ifx \shownote     \undefined \def \shownote      #1{#1}          \fi
\ifx \showarticletitle \undefined \def \showarticletitle #1{#1}   \fi
\ifx \showURL      \undefined \def \showURL       {\relax}        \fi
\providecommand\bibfield[2]{#2}
\providecommand\bibinfo[2]{#2}
\providecommand\natexlab[1]{#1}
\providecommand\showeprint[2][]{arXiv:#2}

\bibitem[\protect\citeauthoryear{Bast, Delling, Goldberg, M{\"u}ller-Hannemann,
  Pajor, Sanders, Wagner, and Werneck}{Bast et~al\mbox{.}}{2016}]%
        {bast2016route}
\bibfield{author}{\bibinfo{person}{Hannah Bast}, \bibinfo{person}{Daniel
  Delling}, \bibinfo{person}{Andrew Goldberg}, \bibinfo{person}{Matthias
  M{\"u}ller-Hannemann}, \bibinfo{person}{Thomas Pajor}, \bibinfo{person}{Peter
  Sanders}, \bibinfo{person}{Dorothea Wagner}, {and} \bibinfo{person}{Renato~F
  Werneck}.} \bibinfo{year}{2016}\natexlab{}.
\newblock \showarticletitle{Route planning in transportation networks}.
\newblock In \bibinfo{booktitle}{\emph{Algorithm engineering}}.
  \bibinfo{publisher}{Springer}, \bibinfo{pages}{19--80}.
\newblock


\bibitem[\protect\citeauthoryear{Ben-Akiva, Bierlaire, Koutsopoulos, and
  Mishalani}{Ben-Akiva et~al\mbox{.}}{1998}]%
        {ben1998dynamit}
\bibfield{author}{\bibinfo{person}{Moshe Ben-Akiva}, \bibinfo{person}{Michel
  Bierlaire}, \bibinfo{person}{Haris Koutsopoulos}, {and} \bibinfo{person}{Rabi
  Mishalani}.} \bibinfo{year}{1998}\natexlab{}.
\newblock \showarticletitle{DynaMIT: A simulation-based system for traffic
  prediction}. In \bibinfo{booktitle}{\emph{DACCORD Short Term Forecasting
  Workshop}}. Delft, The Netherlands, \bibinfo{pages}{1--12}.
\newblock


\bibitem[\protect\citeauthoryear{Bruna, Zaremba, Szlam, and Lecun}{Bruna
  et~al\mbox{.}}{2014}]%
        {bruna2014spectral}
\bibfield{author}{\bibinfo{person}{Joan Bruna}, \bibinfo{person}{Wojciech
  Zaremba}, \bibinfo{person}{Arthur Szlam}, {and} \bibinfo{person}{Yann
  Lecun}.} \bibinfo{year}{2014}\natexlab{}.
\newblock \showarticletitle{Spectral networks and locally connected networks on
  graphs}. In \bibinfo{booktitle}{\emph{Proceedings of the 2nd International
  Conference on Learning Representations (ICLR)}}.
\newblock


\bibitem[\protect\citeauthoryear{Burghout}{Burghout}{2004}]%
        {burghout2004hybrid}
\bibfield{author}{\bibinfo{person}{Wilco Burghout}.}
  \bibinfo{year}{2004}\natexlab{}.
\newblock \emph{\bibinfo{title}{Hybrid microscopic-mesoscopic traffic
  simulation modelling}}.
\newblock \bibinfo{thesistype}{Ph.D. Dissertation}. \bibinfo{school}{PhD
  thesis, Dept of Infrastracture, Royal Institute of Technology, Stockholm,
  Sweden}.
\newblock


\bibitem[\protect\citeauthoryear{Clevert, Unterthiner, and Hochreiter}{Clevert
  et~al\mbox{.}}{2016}]%
        {clevert2015fast}
\bibfield{author}{\bibinfo{person}{Djork-Arn{\'e} Clevert},
  \bibinfo{person}{Thomas Unterthiner}, {and} \bibinfo{person}{Sepp
  Hochreiter}.} \bibinfo{year}{2016}\natexlab{}.
\newblock \showarticletitle{Fast and accurate deep network learning by
  exponential linear units (ELUs)}. In \bibinfo{booktitle}{\emph{Proceedings of
  the 4th International Conference on Learning Representations (ICLR)}}.
\newblock


\bibitem[\protect\citeauthoryear{Dailey and Ted}{Dailey and Ted}{2006}]%
        {dailey2006use}
\bibfield{author}{\bibinfo{person}{Daniel~J Dailey} {and}
  \bibinfo{person}{Trepanier Ted}.} \bibinfo{year}{2006}\natexlab{}.
\newblock \bibinfo{booktitle}{\emph{The use of weather data to predict
  non-recurring traffic congestion}}.
\newblock \bibinfo{type}{{T}echnical {R}eport}. \bibinfo{institution}{Technical
  report to Washington State Transportation Commission, Washington State
  Department of Transportation, University of Washington TransNow, and Federal
  Highway Administration}.
\newblock


\bibitem[\protect\citeauthoryear{Defferrard, Bresson, and
  Vandergheynst}{Defferrard et~al\mbox{.}}{2016}]%
        {defferrard2016convolutional}
\bibfield{author}{\bibinfo{person}{Micha{\"e}l Defferrard},
  \bibinfo{person}{Xavier Bresson}, {and} \bibinfo{person}{Pierre
  Vandergheynst}.} \bibinfo{year}{2016}\natexlab{}.
\newblock \showarticletitle{Convolutional neural networks on graphs with fast
  localized spectral filtering}. In \bibinfo{booktitle}{\emph{Advances in
  neural information processing systems}}. \bibinfo{pages}{3844--3852}.
\newblock


\bibitem[\protect\citeauthoryear{Fang, Zhang, Meng, Xiang, and Pan}{Fang
  et~al\mbox{.}}{2019}]%
        {fang2019gstnet}
\bibfield{author}{\bibinfo{person}{Shen Fang}, \bibinfo{person}{Qi Zhang},
  \bibinfo{person}{Gaofeng Meng}, \bibinfo{person}{Shiming Xiang}, {and}
  \bibinfo{person}{Chunhong Pan}.} \bibinfo{year}{2019}\natexlab{}.
\newblock \showarticletitle{Gstnet: Global spatial-temporal network for traffic
  flow prediction}. In \bibinfo{booktitle}{\emph{Proceedings of the
  Twenty-Eighth International Joint Conference on Artificial Intelligence,
  IJCAI}}. \bibinfo{pages}{10--16}.
\newblock


\bibitem[\protect\citeauthoryear{Goodfellow, Bengio, Courville, and
  Bengio}{Goodfellow et~al\mbox{.}}{2016}]%
        {goodfellow2016deep}
\bibfield{author}{\bibinfo{person}{Ian Goodfellow}, \bibinfo{person}{Yoshua
  Bengio}, \bibinfo{person}{Aaron Courville}, {and} \bibinfo{person}{Yoshua
  Bengio}.} \bibinfo{year}{2016}\natexlab{}.
\newblock \bibinfo{booktitle}{\emph{Deep learning}}. Vol.~\bibinfo{volume}{1}.
\newblock \bibinfo{publisher}{MIT press Cambridge}.
\newblock


\bibitem[\protect\citeauthoryear{He, Shen, Divakaruni, Wynter, and Lawrence}{He
  et~al\mbox{.}}{2013}]%
        {he2013improving}
\bibfield{author}{\bibinfo{person}{Jingrui He}, \bibinfo{person}{Wei Shen},
  \bibinfo{person}{Phani Divakaruni}, \bibinfo{person}{Laura Wynter}, {and}
  \bibinfo{person}{Rick Lawrence}.} \bibinfo{year}{2013}\natexlab{}.
\newblock \showarticletitle{Improving traffic prediction with tweet semantics}.
  In \bibinfo{booktitle}{\emph{Proceedings of the 23rd International Joint
  Conference on Artificial Intelligence (IJCAI)}}. \bibinfo{pages}{1387--1393}.
\newblock


\bibitem[\protect\citeauthoryear{Hoogendoorn and Bovy}{Hoogendoorn and
  Bovy}{2001}]%
        {hoogendoorn2001state}
\bibfield{author}{\bibinfo{person}{Serge~P Hoogendoorn} {and}
  \bibinfo{person}{Piet~HL Bovy}.} \bibinfo{year}{2001}\natexlab{}.
\newblock \showarticletitle{State-of-the-art of vehicular traffic flow
  modelling}.
\newblock \bibinfo{journal}{\emph{Proceedings of the Institution of Mechanical
  Engineers, Part I: Journal of Systems and Control Engineering}}
  \bibinfo{volume}{215}, \bibinfo{number}{4} (\bibinfo{year}{2001}),
  \bibinfo{pages}{283--303}.
\newblock


\bibitem[\protect\citeauthoryear{Karypis and Kumar}{Karypis and Kumar}{1998}]%
        {karypis1998fast}
\bibfield{author}{\bibinfo{person}{George Karypis} {and} \bibinfo{person}{Vipin
  Kumar}.} \bibinfo{year}{1998}\natexlab{}.
\newblock \showarticletitle{A fast and high quality multilevel scheme for
  partitioning irregular graphs}.
\newblock \bibinfo{journal}{\emph{SIAM Journal on scientific Computing}}
  \bibinfo{volume}{20}, \bibinfo{number}{1} (\bibinfo{year}{1998}),
  \bibinfo{pages}{359--392}.
\newblock


\bibitem[\protect\citeauthoryear{Kingma and Ba}{Kingma and Ba}{2015}]%
        {kingma2014adam}
\bibfield{author}{\bibinfo{person}{Diederik~P Kingma} {and}
  \bibinfo{person}{Jimmy Ba}.} \bibinfo{year}{2015}\natexlab{}.
\newblock \showarticletitle{Adam: A method for stochastic optimization}. In
  \bibinfo{booktitle}{\emph{Proceedings of the 3rd International Conference on
  Learning Representations (ICLR)}}.
\newblock


\bibitem[\protect\citeauthoryear{Koesdwiady, Soua, and Karray}{Koesdwiady
  et~al\mbox{.}}{2016}]%
        {koesdwiady2016improving}
\bibfield{author}{\bibinfo{person}{Arief Koesdwiady}, \bibinfo{person}{Ridha
  Soua}, {and} \bibinfo{person}{Fakhreddine Karray}.}
  \bibinfo{year}{2016}\natexlab{}.
\newblock \showarticletitle{Improving prediction with weather information in
  connected cars: A deep learning approach}.
\newblock \bibinfo{journal}{\emph{IEEE Transactions on Vehicular Technology}}
  \bibinfo{volume}{65}, \bibinfo{number}{12} (\bibinfo{year}{2016}),
  \bibinfo{pages}{9508--9517}.
\newblock


\bibitem[\protect\citeauthoryear{Li, Yu, Shahabi, and Liu}{Li
  et~al\mbox{.}}{2018}]%
        {li2018diffusion}
\bibfield{author}{\bibinfo{person}{Yaguang Li}, \bibinfo{person}{Rose Yu},
  \bibinfo{person}{Cyrus Shahabi}, {and} \bibinfo{person}{Yan Liu}.}
  \bibinfo{year}{2018}\natexlab{}.
\newblock \showarticletitle{Diffusion convolutional recurrent neural network:
  Data-driven traffic forecasting}. In \bibinfo{booktitle}{\emph{Proceedings of
  the 6th International Conference on Learning Representations (ICLR)}}.
\newblock


\bibitem[\protect\citeauthoryear{Liao, Zhang, Wu, McIlwraith, Chen, Yang, Guo,
  and Wu}{Liao et~al\mbox{.}}{2018}]%
        {liao2018deep}
\bibfield{author}{\bibinfo{person}{Binbing Liao}, \bibinfo{person}{Jingqing
  Zhang}, \bibinfo{person}{Chao Wu}, \bibinfo{person}{Douglas McIlwraith},
  \bibinfo{person}{Tong Chen}, \bibinfo{person}{Shengwen Yang},
  \bibinfo{person}{Yike Guo}, {and} \bibinfo{person}{Fei Wu}.}
  \bibinfo{year}{2018}\natexlab{}.
\newblock \showarticletitle{Deep Sequence Learning with Auxiliary Information
  for Traffic Prediction}. In \bibinfo{booktitle}{\emph{Proceedings of the 24th
  ACM SIGKDD International Conference on Knowledge Discovery and Data Mining}}.
  ACM.
\newblock


\bibitem[\protect\citeauthoryear{Lou, Zhang, Zheng, Xie, Wang, and Huang}{Lou
  et~al\mbox{.}}{2009}]%
        {lou2009map}
\bibfield{author}{\bibinfo{person}{Yin Lou}, \bibinfo{person}{Chengyang Zhang},
  \bibinfo{person}{Yu Zheng}, \bibinfo{person}{Xing Xie}, \bibinfo{person}{Wei
  Wang}, {and} \bibinfo{person}{Yan Huang}.} \bibinfo{year}{2009}\natexlab{}.
\newblock \showarticletitle{Map-matching for low-sampling-rate GPS
  trajectories}. In \bibinfo{booktitle}{\emph{Proceedings of the 17th ACM
  SIGSPATIAL international conference on advances in geographic information
  systems}}. \bibinfo{pages}{352--361}.
\newblock


\bibitem[\protect\citeauthoryear{Lv, Duan, Kang, Li, Wang, et~al\mbox{.}}{Lv
  et~al\mbox{.}}{2015}]%
        {lv2015traffic}
\bibfield{author}{\bibinfo{person}{Yisheng Lv}, \bibinfo{person}{Yanjie Duan},
  \bibinfo{person}{Wenwen Kang}, \bibinfo{person}{Zhengxi Li},
  \bibinfo{person}{Fei-Yue Wang}, {et~al\mbox{.}}}
  \bibinfo{year}{2015}\natexlab{}.
\newblock \showarticletitle{Traffic flow prediction with big data: A deep
  learning approach.}
\newblock \bibinfo{journal}{\emph{IEEE Trans. Intelligent Transportation
  Systems}} \bibinfo{volume}{16}, \bibinfo{number}{2} (\bibinfo{year}{2015}),
  \bibinfo{pages}{865--873}.
\newblock


\bibitem[\protect\citeauthoryear{Pascale and Nicoli}{Pascale and
  Nicoli}{2011}]%
        {pascale2011adaptive}
\bibfield{author}{\bibinfo{person}{Alessandra Pascale} {and}
  \bibinfo{person}{Monica Nicoli}.} \bibinfo{year}{2011}\natexlab{}.
\newblock \showarticletitle{Adaptive Bayesian network for traffic flow
  prediction}. In \bibinfo{booktitle}{\emph{2011 IEEE Statistical Signal
  Processing Workshop (SSP)}}. IEEE, \bibinfo{pages}{177--180}.
\newblock


\bibitem[\protect\citeauthoryear{Tsymbal}{Tsymbal}{2004}]%
        {tsymbal2004problem}
\bibfield{author}{\bibinfo{person}{Alexey Tsymbal}.}
  \bibinfo{year}{2004}\natexlab{}.
\newblock \showarticletitle{The problem of concept drift: definitions and
  related work}.
\newblock \bibinfo{journal}{\emph{Computer Science Department, Trinity College
  Dublin}} \bibinfo{volume}{106}, \bibinfo{number}{2} (\bibinfo{year}{2004}),
  \bibinfo{pages}{58}.
\newblock


\bibitem[\protect\citeauthoryear{Vlahogianni}{Vlahogianni}{2015}]%
        {vlahogianni2015computational}
\bibfield{author}{\bibinfo{person}{Eleni~I Vlahogianni}.}
  \bibinfo{year}{2015}\natexlab{}.
\newblock \showarticletitle{Computational intelligence and optimization for
  transportation big data: challenges and opportunities}.
\newblock In \bibinfo{booktitle}{\emph{Engineering and Applied Sciences
  Optimization}}. \bibinfo{publisher}{Springer}, \bibinfo{pages}{107--128}.
\newblock


\bibitem[\protect\citeauthoryear{Von~Luxburg}{Von~Luxburg}{2007}]%
        {von2007tutorial}
\bibfield{author}{\bibinfo{person}{Ulrike Von~Luxburg}.}
  \bibinfo{year}{2007}\natexlab{}.
\newblock \showarticletitle{A tutorial on spectral clustering}.
\newblock \bibinfo{journal}{\emph{Statistics and computing}}
  \bibinfo{volume}{17}, \bibinfo{number}{4} (\bibinfo{year}{2007}),
  \bibinfo{pages}{395--416}.
\newblock


\bibitem[\protect\citeauthoryear{Wei, Zheng, Yao, and Li}{Wei
  et~al\mbox{.}}{2018}]%
        {guanjiezhengkdd18}
\bibfield{author}{\bibinfo{person}{Hua Wei}, \bibinfo{person}{Guanjie Zheng},
  \bibinfo{person}{Huaxiu Yao}, {and} \bibinfo{person}{Zhenhui Li}.}
  \bibinfo{year}{2018}\natexlab{}.
\newblock \showarticletitle{Intellilight: A reinforcement learning approach for
  intelligent traffic light control}. In \bibinfo{booktitle}{\emph{Proceedings
  of the 24th {ACM} {SIGKDD} International Conference on Knowledge Discovery
  {\&} Data Mining, {KDD} 2018, London, UK, August 19-23, 2018}}.
  \bibinfo{pages}{2496--2505}.
\newblock


\bibitem[\protect\citeauthoryear{Yu, Yin, and Zhu}{Yu et~al\mbox{.}}{2018}]%
        {yu2018spatio}
\bibfield{author}{\bibinfo{person}{Bing Yu}, \bibinfo{person}{Haoteng Yin},
  {and} \bibinfo{person}{Zhanxing Zhu}.} \bibinfo{year}{2018}\natexlab{}.
\newblock \showarticletitle{Spatio-Temporal Graph Convolutional Neural Network:
  A Deep Learning Framework for Traffic Forecasting}. In
  \bibinfo{booktitle}{\emph{Proceedings of the 27th International Joint
  Conference on Artificial Intelligence (IJCAI)}}.
\newblock


\bibitem[\protect\citeauthoryear{Zhang, Wang, Wang, Lin, Xu, and Chen}{Zhang
  et~al\mbox{.}}{2011}]%
        {zhang2011data}
\bibfield{author}{\bibinfo{person}{Junping Zhang}, \bibinfo{person}{Fei-Yue
  Wang}, \bibinfo{person}{Kunfeng Wang}, \bibinfo{person}{Wei-Hua Lin},
  \bibinfo{person}{Xin Xu}, {and} \bibinfo{person}{Cheng Chen}.}
  \bibinfo{year}{2011}\natexlab{}.
\newblock \showarticletitle{Data-driven intelligent transportation systems: A
  survey}.
\newblock \bibinfo{journal}{\emph{IEEE Transactions on Intelligent
  Transportation Systems}} \bibinfo{volume}{12}, \bibinfo{number}{4}
  (\bibinfo{year}{2011}), \bibinfo{pages}{1624--1639}.
\newblock


\bibitem[\protect\citeauthoryear{Zheng, Fan, Wen, Chen, Wang, and Li}{Zheng
  et~al\mbox{.}}{2019}]%
        {zheng2019deepstd}
\bibfield{author}{\bibinfo{person}{Chuanpan Zheng}, \bibinfo{person}{Xiaoliang
  Fan}, \bibinfo{person}{Chenglu Wen}, \bibinfo{person}{Longbiao Chen},
  \bibinfo{person}{Cheng Wang}, {and} \bibinfo{person}{Jonathan Li}.}
  \bibinfo{year}{2019}\natexlab{}.
\newblock \showarticletitle{DeepSTD: Mining spatio-temporal disturbances of
  multiple context factors for citywide traffic flow prediction}.
\newblock \bibinfo{journal}{\emph{IEEE Transactions on Intelligent
  Transportation Systems}} (\bibinfo{year}{2019}).
\newblock


\end{thebibliography}

\end{document}